%% file: main.tex
\documentclass{article}

\input{math_commands}

\usepackage{microtype}
\usepackage{graphicx}
\usepackage{subfigure}
\usepackage{booktabs} % for professional tables

\usepackage{hyperref}

\usepackage[accepted]{icml2025}

\usepackage{amsmath}
\usepackage{amssymb}
\usepackage{mathtools}
\usepackage{amsthm}

\usepackage[capitalize,noabbrev]{cleveref}

\newcommand{\bz}{\boldsymbol{z}}
\newcommand{\bx}{\boldsymbol{x}}
\newcommand{\bu}{\textbf{u}}

\newcommand{\tr}{\text{tr}} % trace

\newcommand{\bQ}{\textbf{Q}}
\newcommand{\bK}{\textbf{K}}
\newcommand{\bX}{\textbf{X}}
\newcommand{\Real}{\mathbb{R}}
\newcommand{\bI}{\textbf{I}}
\newcommand{\bZ}{\textbf{Z}}

\newcommand{\bV}{\textbf{V}}
\newcommand{\bS}{\textbf{S}}
\newcommand{\bA}{\textbf{A}}
\newcommand{\K}{\textbf{K}}
\newcommand{\f}{\textbf{f}}
\newcommand{\y}{\textbf{y}}
\newcommand{\bk}{\textbf{k}}
\newcommand{\m}{\textbf{m}}
\newcommand{\bfmu}{\boldsymbol{\mu}}
\newcommand{\bLambda}{\boldsymbol{\Lambda}}

\newcommand{\comm}[1]{}

\theoremstyle{plain}
\newtheorem{theorem}{Theorem}[section]
\newtheorem{proposition}[theorem]{Proposition}
\newtheorem{lemma}[theorem]{Lemma}

\theoremstyle{definition}

\theoremstyle{remark}
\newtheorem{remark}[theorem]{Remark}

\usepackage[textsize=tiny]{todonotes}
\usepackage{cancel}

\icmltitlerunning{New Bounds for Sparse Variational Gaussian Processes}

\begin{document}

\twocolumn[
\icmltitle{New Bounds for Sparse Variational Gaussian Processes}



\icmlsetsymbol{equal}{*}

\begin{icmlauthorlist}
\icmlauthor{Michalis K. Titsias}{comp}
\end{icmlauthorlist}

\icmlaffiliation{comp}{Google DeepMind}

\icmlcorrespondingauthor{Michalis K. Titsias}{mtitsias@google.com}

\icmlkeywords{Sparse Gaussian Processes, Machine Learning, ICML}

\vskip 0.3in
]



\printAffiliationsAndNotice{}  % leave blank if no need to mention equal contribution

\begin{abstract}
Sparse variational Gaussian processes (GPs) construct tractable posterior approximations to GP  models. At the core of these methods is the assumption that the true posterior distribution over training function values $\f$ and inducing variables $\bu$ is approximated by a variational distribution that incorporates the conditional GP prior $p(\f | \bu)$ in its factorization. While  this assumption is considered as fundamental, 
%Imposing this conditional prior in the approximation is believed to be a fundamental  requirement to obtain scalable GPs. 
we show that for model training we can relax it through the use of a more general variational distribution $q(\f | \bu)$ that depends on $N$ extra parameters, where  $N$ is the number of training examples. In GP regression, we can analytically optimize
the evidence lower bound over the extra parameters and express a tractable collapsed bound that is tighter than the previous bound. The new bound is also amenable to stochastic %gradient 
optimization and its implementation requires minor modifications to existing sparse GP code. 
Further, we also describe extensions to non-Gaussian likelihoods. 
On several %regression 
datasets we demonstrate that our method can reduce  bias when learning the %model
hyperparameters and can lead to better predictive performance. 
%such as applications to GP Poisson regression. 
\end{abstract}

\input{sec_intro}

\input{sec_background}

\input{sec_methods}
\input{sec_related}
\input{sec_experiments}

\input{sec_conclusion}

%\section*{Accessibility}
%Authors are kindly asked to make %their submissions as accessible as %possible for everyone including %people with disabilities and sensory %or neurological differences.
%Tips of how to achieve this and what to pay attention to will be provided on the conference website \url{http://icml.cc/}.

%\section*{Software and Data}
%If a paper is accepted, we strongly encourage the publication of software and data with the
%camera-ready version of the paper whenever appropriate. This can be
%done by including a URL in the camera-ready copy. However, \textbf{do not}
%include URLs that reveal your institution or identity in your
%submission for review. Instead, provide an anonymous URL or upload
%the material as ``Supplementary Material'' into the CMT reviewing
%system. Note that reviewers are not required to look at this material
%when writing their review.

%% Acknowledgements should only appear in the accepted version.
\section*{Acknowledgements}

I am grateful to the reviewers for their comments. Also, I wish to thank Jiaxin Shi and Francisco Ruiz for their invaluable advice and generous help during this project.

\section*{Impact Statement}

This paper presents work whose goal is to advance the field of 
Machine Learning. There are many potential societal consequences 
of our work, none which we feel must be specifically highlighted here.

%If a paper is accepted, the final camera-ready version can (and
%probably should) include acknowledgements. In this case, please
%place such acknowledgements in an unnumbered section at the
%end of the paper. Typically, this will include thanks to reviewers
%who gave useful comments, to colleagues who contributed to the ideas,
%and to funding agencies and corporate sponsors that provided financial support.

% references without citing it in the main text, use \nocite \nocite{langley00}

\bibliography{library}
\bibliographystyle{icml2025}

%%%%%%%%%%%%%%%%%%%%%%%%%%%%%%%%%%%%%%%%%%%%%%%%%%%%%%%%%%%%%%%%%%%%%%%%%%%%%%%
%%%%%%%%%%%%%%%%%%%%%%%%%%%%%%%%%%%%%%%%%%%%%%%%%%%%%%%%%%%%%%%%%%%%%%%%%%%%%%%
% APPENDIX
%%%%%%%%%%%%%%%%%%%%%%%%%%%%%%%%%%%%%%%%%%%%%%%%%%%%%%%%%%%%%%%%%%%%%%%%%%%%%%%
%%%%%%%%%%%%%%%%%%%%%%%%%%%%%%%%%%%%%%%%%%%%%%%%%%%%%%%%%%%%%%%%%%%%%%%%%%%%%%%
\newpage
\appendix
\onecolumn
\input{appendix}

%You can have as much text here as you want. The main body must be at most $8$ pages long. For the final version, one more page can be added. If you want, you can use an appendix like this one, even using the one-column format.
%%%%%%%%%%%%%%%%%%%%%%%%%%%%%%%%%%%%%%%%%%%%%%%%%%%%%%%%%%%%%%%%%%%%%%%%%%%%%%%
%%%%%%%%%%%%%%%%%%%%%%%%%%%%%%%%%%%%%%%%%%%%%%%%%%%%%%%%%%%%%%%%%%%%%%%%%%%%%%%

\end{document}

%% file: math_commands.tex
\usepackage{amsmath,amsfonts,bm}

\def\eqref#1{equation~\ref{#1}}
\def\1{\bm{1}}

\DeclareMathAlphabet{\mathsfit}{\encodingdefault}{\sfdefault}{m}{sl}
\SetMathAlphabet{\mathsfit}{bold}{\encodingdefault}{\sfdefault}{bx}{n}

\newcommand{\E}{\mathbb{E}}

%% file: sec_intro.tex
\section{Introduction}
\label{sec:intro}

Gaussian processes (GPs) are  nonparametric models for learning %unknown 
functions using Bayesian learning. Thanks to their flexibility and ability to quantify uncertainty, GPs have found many applications in 
 machine learning \cite{rasmussen2006gaussian}, spatial  modeling \cite{Cressie1993}, computer experiments \cite{OHagan78,gramacy2020surrogates},  Bayesian optimization \cite{Jones1998EfficientGO,garnett_bayesoptbook_2023}, robotics and control \cite{deisenroth2011pilco}, unsupervised learning 
\citep{lawrence2005probabilistic} and  others. 

Despite the numerous applications, GPs  suffer from $\mathcal{O}(N^3)$ time cost  and $\mathcal{O}(N^2)$ storage where  $N$ is the number of training examples. This has originated a large body of research on scalable or sparse GP methods
expanded in several decades; see e.g., Chapter 8 in \citet{rasmussen2006gaussian}  
for an early review and  \citet{heaton2018, LiuetalGPreview20, leibfried2022tutorialsgp} for recent treatments. 
An important class of methods 
 bases an approximation on a small set of $M \ll  N$ 
inducing points 
\citep{csato-opper-02,lawrence-seeger-herbrich-01, seeger-williams-lawrence-03a,Snelson2006, candela-rasmussen-05,Banerjee2008,Finley2009,titsias2009variational, hensman2013gaussian,Buietal2017,burt2020convergence}
that reduce the time complexity to $\mathcal{O}(N M^2)$ and 
the storage to $\mathcal{O}(N M)$. 
% GPs have also been unfavourably compared to deep learning models for lacking representation learning capabilities.

Among inducing point methods, the sparse variational Gaussian process (SVGP), introduced for standard regression \cite{titsias2009variational}, 
applies variational inference to obtain a posterior approximation and selects hyperparameters and inducing points by maximizing an evidence lower bound. Unlike 
the prior approximation framework 
\cite{candela-rasmussen-05}, SVGP
leaves the GP prior unchanged and instead it reduces the cost to $\mathcal{O}(N M^2)$ 
by imposing a special structure on the variational 
distribution. This framework has been extended
to stochastic gradient optimization
\cite{hensman2013gaussian} and non-Gaussian likelihoods \cite{Chai12,hensman2015scalable,lloyd15,Dezfouli15,Sheth15}. Also, it has been explained  as KL minimization between stochastic processes \cite{Matthews2015OnSV}.

An important aspect of the SVGP method is that it uses a special form for the variational distribution. It approximates the exact posterior distribution $p(\f, \bu | \y)$ over the training function values $\f$ and the
inducing variables $\bu$ 
% iven data $\y$ 
(see \Cref{sec:background} for precise definitions)
by a variational distribution of the form $q(\f, \bu) = p(\f | \bu) q(\bu)$, where
$q(\bu)$ is some optimizable  distribution over the inducing variables, while
$p(\f | \bu)$ is the conditional GP prior. 
This special form of the variational approximation seems to be fundamental, and it has been applied also to more complex GP models, such as those with multiple outputs 
\cite{alvarez10a,NguyenBonilla14,fariba19}, uncertain inputs \cite{titsias10a,damianou16a} and multiple layers \cite{damianou13a,salimbeni2017doubly}. However, an open question regarding
the SVGP framework is whether this particular form of variational distribution
is really necessary to obtain scalable computations. The answer we give in this paper is that “it is not", since at least for training a GP model it can be relaxed. 

To this end, we derive new variational bounds for training 
sparse GP regression models by replacing 
$p(\f |\bu)$ in the variational distribution with a more general conditional distribution $q(\f | \bu)$. This $q(\f | \bu)$ depends on $N$ additional parameters (on top of the parameters of $p(\f | \bu)$), i.e., as many as the training examples, and it is constructed to enable better covariance approximation of the underlying true factor $p(\f | \bu, \y)$. We show how to analytically optimize over the $N$ parameters and obtain a better posterior approximation together with a tighter 
collapsed evidence lower bound. 
The new bound is also amenable to stochastic gradient optimization, and its
simple form suggests that  it can be implemented with minor modifications to existing sparse GP code. We also describe extensions of the method to non-Gaussian likelihoods. Furthermore, we point out the 
concurrent work of  \citet{bui2025tighter} who derived similar sparse GP approximations
and variational training objectives by using the same form for the $q(\f | \bu)$ distribution. 

The remainder of the paper is as follows. \Cref{sec:background} provides 
an overview of GPs and the variational approach to sparse GPs using inducing points. \Cref{sec:proposedmethod} 
derives the new evidence lower bounds for training. \Cref{sec:relatedwork} discusses connections with previous works. %especially with the work of \citet{artemevburt2021cglb} 
%and the method of \citet{Buietal2017}. 
\Cref{sec:experiments} presents experiments using several %regression 
datasets showing that the new bounds can reduce underfitting bias and can lead to better predictive performance. \Cref{sec:conclusions} 
concludes with a discussion and suggestions for future work.

%% file: sec_background.tex
\section{Background
\label{sec:background}
}

A GP is a distribution over functions specified by a mean function $m(\bx)$ and a covariance or kernel function $k(\bx, \bx')$, where the kernel function is parametrized by $\theta$. 
By assuming that $m(\bx) = 0$ we denote a GP draw as 
$$
f(\bx) \sim \mathcal{GP}(0, k(\bx, \bx')). 
$$
For a finite set of inputs 
$\bX = \{\bx_n\}_{n=1}^N$ the 
distribution over the function values $\f
= \{f_n \}_{n=1}^N$ (stored as $N \times 1$ vector with $f_n := f(\bx_n)$) is 
the multivariate Gaussian $p(\f) = \mathcal{N}(\f | {\bf 0}, \bK_{\f \f} )$ where the $N \times N$ covariance 
matrix $\bK_{\f \f}$ has entries $[\bK_{\f \f}]_{i j} = k(\bx_i, \bx_j)$. 

We consider standard GP regression
where we are given a set of training inputs
$\bX$ and corresponding noisy outputs 
$\y = \{y_n\}_{n=1}^N$ where 
$y_n \in \Real$. Conditionally on the latent values $\f$, these outputs follow a factorized
Gaussian likelihood, $p(\y | \f) = \prod_{n=1}^N \mathcal{N}(y_n | f_n, \sigma^2) = \mathcal{N}(\y | \f, \sigma^2 I)$. The joint distribution over outputs $\y$ 
and latent values $\f$ is
\begin{equation}
p(\y | \f) p(\f) = \mathcal{N}(\y | \f, \sigma^2 I) \mathcal{N}(\f | {\bf 0}, \bK_{\f \f} ).
\end{equation}
To learn the hyperparameters 
$(\theta, \sigma^2)$ we can maximize the
log marginal likelihood which is analytically available,
\begin{equation}
\log p(\y) = \int p(\y | \f) p(\f) d \f = \log \mathcal{N}(\y | {\bf 0 }, \bK_{\f \f}  + \sigma^2 \bI). 
\label{eq:exact_marg_likel}
\end{equation}
After training we can perform predictions at test inputs $\bX_*$
by first computing the posterior 
over the corresponding test function values $\f_*$:
\begin{align}
& p(\f_* | \y) = \int p(\f_* | \f) p(\f | \y) d \f = \\
 & \mathcal{N}(\f_* | \bK_{\f_* \f} (\bK_{\f \f} \! + \! \sigma^2 \bI)^{-1} \f,  \bK_{\f_* \f_*} \! \! - \! \bK_{\f_* \f} (\bK_{\f \f} \! + \! \sigma^2 \bI)^{-1} \bK_{\f \f_*} \! ) \nonumber 
% \label{eq:fstar_giveny}
\end{align} 
and then writing the predictive density as $p(\y_* | \y) = \int \mathcal{N}(\y_* | \f_*, \sigma^2 \bI) p(\f_* | \y) d \f_*$, which is the same as the above Gaussian but with  $\sigma^2 \bI$ added to the covariance. 
 
While the log marginal likelihood and predictive density have closed-form
expressions, they require 
the inversion of $\bK_{\f\f} + \sigma^2 \bI$ which costs $\mathcal{O}(N^3)$ and it is prohibitive for large datasets.  
Next we review methods using inducing 
points and particularly the 
variational approach \cite{titsias2009variational} that our method in \Cref{sec:proposedmethod} improves upon. 

\subsection{Sparse Variational Gaussian Process (SVGP)} 

The idea of inducing points 
 is to base a GP approximation 
on a smaller set of $M \ll N$  function values; see e.g., \citet{csato-opper-02,seeger-williams-lawrence-03a,Snelson2006,candela-rasmussen-05}. \citet{Snelson2006}
introduced pseudo inputs 
by instantiating extra GP function values $\bu = \{f(\bz_m)\}_{m=1}^M$ evaluated at locations $\bZ = \{\bz_m\}_{m=1}^M$ 
that can be optimized freely with gradient-based methods. However, 
the GP prior modification procedure 
\cite{candela-rasmussen-05,Snelson2006} 
does not result in a rigorous  approximation to the GP model. An alternative variational inference method  
\cite{titsias2009variational}, next referred to as
SVGP\footnote{Another common name for this method is Variational Free Energy (VFE); see \citet{Buietal2017,baueretal16,LiuetalGPreview20}.}, does not modify the GP prior but instead it augments the model with extra function values $\bu$: 
\begin{align}
& p(\y, \f, \bu) 
 \! = \! p(\y | \f) 
p(\f | \bu) p(\bu) \ \ \ \ \ \ \ \  \ \ \ \  \ \ \text{augmented joint}
\label{eq:augm-joint}
\\ 
& p(\f | \bu)   
\! = \! \mathcal{N}(\f | \K_{\f \f} \K_{\bu \bu}^{-1} \bu, \K_{\f \f} \! - \! \K_{\f \bu} \K_{\bu \bu}^{-1} \K_{\bu  \f}) \ \ \ \text{cond. GP} 
\label{eq:cond-gp-prior}
\\
& p(\bu) \! = \! \mathcal{N}(\bu | {\bf 0}, \K_{\bu \bu})  \ \ \ \ \ \ \ \ \ \ \ \ \ \ \ \ \ \ \ \ \ \ \ \ \text{inducing GP prior}
\label{eq:ind-gp-prior}
\end{align}
where $\K_{\bu \bu}$ is the $M \times M$ covariance matrix on the 
inducing inputs $\bZ$, 
$\K_{\f \bu}$ is the $N \times M$ cross covariance between points in $\bX$ and $\bZ$, while $\bK_{\bu \f} = \K_{\f \bu}^\top$.  SVGP % method 
approximates the exact posterior $p(\f, \bu|\y)$ by a variational 
distribution $q(\f, \bu)$ through the minimization of  $\text{KL}[q(\f, \bu) || p(\f, \bu|\y)]$. A critical assumption is the following 
choice for $q$:
\begin{equation}
q(\f, \bu) 
= p(\f | \bu) q(\bu), 
\label{eq:pfuqu}
\end{equation}
where $q(\bu)$ is an optimizable $M$-dimensional variational %(Gaussian)
distribution,  while $p(\f | \bu)$ is the same conditional GP prior from \Cref{eq:cond-gp-prior}
that appears in the joint 
in (\ref{eq:augm-joint}). The KL minimization is expressed as the maximization of an evidence lower bound (ELBO) on the log marginal likelihood, 
\begin{align} 
\log p(\y) & \geq \int p(\f | \bu) q(\bu) \log \frac{p(\y | \f) \cancel{p(\f | \bu)} p(\bu)}{\cancel{p(\f | \bu)} q(\bu)} d \f d \bu \nonumber \\ 
& = \int q(\bu) \log \frac{\exp\{ \int p(\f | \bu) \log p(\y | \f) d \f\} p(\bu)}{q(\bu)} d \bu. \nonumber 
\end{align}
If we optimize over $q(\bu)$ and obtain the optimal choice $q^*(\bu) \propto \exp\left\{\int p(\f | \bu) \log p(\y | \f) d \f \right\} p(\bu)$, then we can substitute this $q^*(\bu)$ in the last line above and express the 
%tightest possible ELBO, the 
so called collapsed bound, having the general form 
$$
\log p(\y) \! \geq \! \mathcal{F} \! = \! \log \! \int \! \exp\left\{\int p(\f | \bu) \log p(\y | \f) d \f \right\} p(\bu) d \bu 
$$
which for the standard GP regression
model takes the form 
\begin{align}
\mathcal{F} = \underbrace{\log \mathcal{N}(\y | {\bf 0}, \bQ_{\f\f} + \sigma^2 \bI)}_{\text{DTC log lik}} 
- \underbrace{\frac{1}{ 2 \sigma^2} 
\text{tr}\left(  \bK_{\f\f} - \bQ_{\f\f} \right)}_{\text{trace term}},
\label{eq:collapsedbound_old}
\end{align}
where $\bQ_{\f\f} = \bK_{\f\bu}\bK_{\bu\bu}^{-1}\bK_{\bu\f}$ is the M-rank Nystr\'om matrix.
The first term in the bound is the deterministic training conditional (DTC) log likelihood \cite{seeger-williams-lawrence-03a, candela-rasmussen-05} 
%Banerjee2008} 
while the second is a regularization 
term which, since $\text{tr}(  \bK_{\f\f} - \bQ_{\f\f}) \geq 0$,  promotes $\bQ_{\f\f}$ to stay close to  $\bK_{\f\f}$. The inducing points $\bZ$ can be learned as variational parameters by maximizing the bound jointly with the hyperparameters $(\theta, 
\sigma^2)$, which requires 
$\mathcal{O}(N M^2)$ operations per optimization step. \citet{hensman2013gaussian} further reduced the operations to $\mathcal{O}(M^3)$ 
per optimization step by applying stochastic minibatch training for  
maximizing the uncollapsed version of the bound; see  \Cref{sec:stochasticopt}. 

To obtain the form of the GP posterior 
over any test function values $\f_*$ we can first write the exact form 
\begin{equation}
p(\f_* | \y) 
= \int p(\f_* | \f, \bu) p(\f,  \bu | \y)  d \f d \bu, 
\label{eq:exact_posteriorGP2}
\end{equation}
where $p(\f_* | \f, \bu)$  is the conditional GP of $\f_*$  given 
training function values $\f$ and inducing values $\bu$ while 
$p(\f,  \bu | \y)$ is the posterior 
over $(\f, \bu)$ written also 
as 
\begin{equation}
p(\f,  \bu | \y) = p(\f | \bu, \y) p(\bu | \y).
\label{eq:exact_augm_posterior}
\end{equation}
The SVGP method approximates $p(\f,  \bu | \y)$ by $q(\f, \bu)$ and therefore by plugging in this $q$ into (\ref{eq:exact_posteriorGP2}) we obtain
\begin{equation}
q(\f_* | \y) 
\! = \! \! \int p(\f_* | \f, \bu) p(\f |  \bu) q(\bu)  d \f d \bu \! = \! \! 
\int p(\f_* | \bu) q(\bu) d \bu, 
\label{eq:variational_posteriorGP}
\end{equation}
where $p(\f_* | \bu) = \int p(\f_* | \f, \bu) p(\f |  \bu) d \f$ comes from the GP consistency. For completeness, in \Cref{app:detailsSVGP}
we include further details about SVGP such as a derivation of the collapsed bound and the 
%exact 
Gaussian form 
of the optimal 
$q^*(\bu)$. 

We conclude this review of SVGP for regression with a couple of remarks that will be useful next. 

\begin{remark}
%[VFE can underfit]
The approximation becomes exact when 
$\bK_{\f \f} = \bQ_{\f \f}$ and the collapsed bound %from (\ref{eq:collapsedbound_old}) 
matches the log marginal likelihood in 
(\ref{eq:exact_marg_likel}). However, to obtain good approximations we may need sufficiently large number of inducing points \cite{burt2020convergence}.  Otherwise the bound will cause underfitting. 
%For instance, by writing a typical kernel function as the one in ??, as $k(\bx, \bx') = \sigma_f^2 c(\bx, \bx')$ where $c(\bx, \bx')\leq 1$ is the correlation function then regularization trace term is written as
%$
%- \frac{\sigma^2_f}{ 2 \sigma^2} 
%\text{tr}\left(  \bC_{\f \f} - \bC_{\f %\bu} \bC_{\bu \bu}^{-1} \bC_{\bu \f} %\right)
%$
%which causes the signal-to-noise 
% ratio $\frac{\sigma_f^2}{\sigma^2}$ 
%(i.e., amplitude parameter $\sigma_f^2$ over the noise variance $\sigma^2$) to be penalized towards small values. 
For instance, as studied by 
\citet{baueretal16} 
and \citet{titsias2009variational} 
the SVGP bound tends to overestimate the noise variance $\sigma^2$. 
%and simultaneously underestimate signal amplitude $\sigma_f^2$. 
\label{remark1}
\end{remark}

\begin{remark}
%[Approximating $p(\f | \bu, \y)$ by $p(\f | \bu)$]
 SVGP approximates $p(\f | \bu, \y)$ in the exact posterior 
in (\ref{eq:exact_augm_posterior}) 
by the conditional GP 
$p(\f | \bu)$, in the variational posterior in (\ref{eq:pfuqu}), 
while  $q(\bu)$ is treated optimally by %free-form 
KL minimization. If 
$p(\f | \bu, \y) \! = \! p(\f | \bu)$ %(where  recall $\f$ are the training function values and not the full infinite function!)
then $\text{KL}[q(\f, \bu) || p(\f, \bu|\y)]=0$ and the % full infinite 
 approximation becomes exact, meaning $q(\f_* | \y) \! = \! p(\f_* | \y)$ for any $\f_* = f_*(\bX_*)$. 
%In other words, the finite vector of training function values $\f$ is special, since it suffices to approximate $\f$ in order to become accurate over the full infinite process. 
\label{remark2}
\end{remark}

%More generally, if we do not collapse $q(\bu)$ and let $q(\bu) = \mathcal{N}(\bbm_\bu,\bS_\bu)$, where $\bbm_\bu, \bS_\bu$ are trainable parameters, we %obtain the uncollapsed bound suitable for can use the uncollapsed bound formini-batch training and non-Gaussian likelihoods~\citep{hensman2013gaussian,hensman2015scalable}.
%\andriy{Briefly explain why, if there's space.}
% Computing this objective requires $\mathcal{O}(M^2N + M^3)$ operations, which can be further reduced by unbiased estimation of the first term using a mini-batch of training data.

%% file: sec_methods.tex
\section{Proposed Method: Tighter Bounds
\label{sec:proposedmethod}
}

Remark \ref{remark1} suggests that 
it would be useful to tighten the collapsed bound 
%in (\ref{eq:collapsedbound_old}) 
in order to reduce underfitting bias 
and match better exact GP training. 
Remark \ref{remark2} suggests that one way to tighten the bound is to replace %the conditional GP 
$p(\f | \bu)$, in the variational approximation in (\ref{eq:pfuqu}), with another distribution 
that can better approximate 
$p(\f | \bu, \y)$. Next we develop a method that does this while keeping the cost unchanged. 

Let us write the 
exact form of $p(\f | \bu, \y)$. By noting that this quantity is the exact posterior over $\f$ in a GP regression model with joint $p(\y | \f) p(\f | \bu)$ 
%(where $p(\f | \bu)$ is now the effective GP prior) 
we conclude that this %posterior 
is 
$$
p(\f | \bu, \y) = \mathcal{N}\left(\f| \m(\y,\bu), 
(\widetilde{\bK}_{\f \f}^{-1} + \frac{1}{\sigma^2} \bI)^{-1} \right),
$$
where $\m(\y,\bu)  = \E[\f | \bu] + \widetilde{\bK}_{\f \f} (\widetilde{\bK}_{\f \f} + \sigma^2 \bI)^{-1} (\y - \E[\f | \bu]) $
with $\E[\f | \bu] = \bK_{\f \bu} \bK_{\bu \bu}^{-1} \bu$ and $\widetilde{\bK}_{\f \f} = \bK_{\f \f} - \bQ_{\f \f}$. Note that under this notation, 
$p(\f | \bu) = \mathcal{N}(\f | \E[\f | \bu], \widetilde{\bK}_{\f \f})$. We will construct a new $q(\f | \bu)$ 
that keeps the same mean $\E[\f | \bu]$ 
as $p(\f | \bu)$ but it replaces $\widetilde{\bK}_{\f \f}$ with a closer approximation to the 
% exact 
covariance 
% matrix 
$(\widetilde{\bK}_{\f \f}^{-1} + \frac{1}{\sigma^2} \bI)^{-1}$ of $p(\f | \bu, \y)$. We first 
write this %latter 
matrix as 
\begin{equation}
(\widetilde{\bK}_{\f \f}^{-1} + \frac{1}{\sigma^2} \bI)^{-1}
= \widetilde{\bK}_{\f \f}^{\frac{1}{2}}
( \bI + \frac{1}{\sigma^2} \widetilde{\bK}_{\f \f})^{-1} 
\widetilde{\bK}_{\f \f}^{\frac{1}{2}}.
\label{eq:exact_cov_pfuy}
\end{equation}
Then we approximate the inverse 
$( \bI + \frac{1}{\sigma^2} \widetilde{\bK}_{\f \f})^{-1}$ by a diagonal matrix $\bV = \text{diag}(v_1, \ldots,v_N)$ of $N$ variational parameters $v_i > 0$. In other words,  in the initial $q(\f, \bu) = p(\f|\bu)q(\bu)$ we will replace $p(\f|\bu)$ by 
\begin{equation}
q(\f|\bu) = \mathcal{N}(\f | \bK_{\f \bu} \bK_{\bu \bu}^{-1} \bu, (\bK_{\f \f} - \bQ_{\f \f})^{\frac{1}{2}} \bV
(\bK_{\f \f} - \bQ_{\f \f})^{\frac{1}{2}}).
\label{eq:qfu}
\end{equation}
The ELBO now is written as 
\begin{align} 
 & \int q(\f | \bu) q(\bu) \log \frac{p(\y | \f) p(\f | \bu) p(\bu)}{q(\f | \bu) q(\bu)} d \f d \bu = \nonumber \\ 
& \int \! \!  q(\bu) \! \left\{ \! \log \frac{e^{\E_{q(\f | \bu)}[\log p(\y | \f)]} p(\bu)}{q(\bu)} \! - \! \text{KL}[q(\f | \bu) || p(\f | \bu)] 
\! \right\} \! d \bu \nonumber 
\end{align}
and the challenge is to see whether 
$\text{KL}[q(\f | \bu) || p(\f | \bu)]$ 
and $\E_{q(\f | \bu)}[\log p(\y | \f)]$ 
are computable in $\mathcal{O}(N M^2)$ time. 
We have the following results (proofs are in  \Cref{app:detailsNewbounds}).
\begin{lemma}
\label{lem:KLqfupfu}
\emph{$\text{KL}[q(\f | \bu) || p(\f | \bu)] 
= \frac{1}{2} \sum_{i=1}^N (v_i - \log v_i - 1)$}.
\end{lemma}
\begin{lemma} 
\label{lem:Expqfu_loglik}
Let us denote the diagonal elements of \emph{$\bK_{\f \f} - \bQ_{\f \f}$} as 
\emph{$k_{ii} - q_{ii}$} for \emph{$i=1,\ldots,N$}. Then  
\emph{\begin{align}
& \E_{q(\f | \bu)}[\log p(\y | \f)] \nonumber \\ 
& \! = \! \log \mathcal{N}(\y | \bK_{\f \bu}
\bK_{\bu \bu }^{-1} \bu, \sigma^2 \bI)
- \frac{1}{2 \sigma^2} 
\sum_{i=1}^N v_i (k_{ii} - q_{ii}). 
\end{align}}
\end{lemma}
By combining the two lemmas the full bound is written as 
\begin{align} 
& \int \! \!  q(\bu) \log \frac{  \mathcal{N}(\y | \bK_{\f \bu}
\bK_{\bu \bu }^{-1} \bu, \sigma^2 \bI) p(\bu)}{q(\bu)}  d \bu \nonumber \\
& - \frac{1}{2} 
\sum_{i=1}^N \left\{  v_i \left(1 + \frac{k_{ii} - q_{ii}}{\sigma^2}\right) - \log v_i -1 \right\}.  
\label{eq:newcollapsedbound_with_vis}
\end{align}
\begin{proposition}%[new collapsed bound]
Maximizing the bound in (\ref{eq:newcollapsedbound_with_vis}) with respect to \emph{$q(\bu)$}
and each \emph{$v_i$} results in the 
optimal settings \emph{$q^*(\bu) \propto  \mathcal{N}(\y | \bK_{\f \bu}
\bK_{\bu \bu }^{-1} \bu, \sigma^2 \bI) p(\bu)$} and  
\emph{$v_i^* = \left(1 + \frac{k_{ii} - q_{ii}}{\sigma^2} \right)^{-1}$}. By substituting these values 
back to (\ref{eq:newcollapsedbound_with_vis}) we obtain
\emph{\begin{equation} 
\mathcal{F}_{new} \! = \! \log  \mathcal{N}(\y |{\bf 0},   \bQ_{\f \f} + \sigma^2 \bI) 
 - \frac{1}{2}  
\sum_{i=1}^N \log \left(\! 1 + \frac{k_{ii} - q_{ii}}{\sigma^2} \! \right).   
\label{eq:newcollapsedbound}
\end{equation}
}
\label{prop:newbound}
\end{proposition}
The first term is the 
DTC log likelihood as in the original bound in (\ref{eq:collapsedbound_old}),  
but the regularization term 
makes the bound tighter, 
i.e., $\log p(\y) \geq \mathcal{F}_{new} \geq \mathcal{F}$, due to the inequality $\log(a + 1) \leq a$. Also since $\log(a + 1) < a$ for all $a>0$, if $\bK_{\f \f} \neq \bQ_{\f \f}$ 
(so there is at least one $k_{ii} - q_{ii} > 0$), then $\mathcal{F}_{new} > \mathcal{F}$. This means that $\mathcal{F}_{new}$ is strictly better than $\mathcal{F}$ unless both bounds match exactly the log marginal likelihood. 

Clearly, $\mathcal{F}_{new}$ has $\mathcal{O}(N M^2)$ cost and its implementation requires a minor modification to the initial bound. The optimal $q^*(\bu)$
is the same as in the initial SVGP method, while 
an interpretation of the optimal $v_i^*$ values
is the following.  
%\begin{remark}
%Recall that $\log(a + 1) < a$ for $ \forall a>0$. Thus, if $\bK_{\f \f} \neq \bQ_{\f \f}$ (so there is at least one $k_{ii} - q_{ii} > 0$),  $\mathcal{F}_{new} > \mathcal{F}$ which means that $\mathcal{F}_{new}$ is strictly better than $\mathcal{F}$ unless both bounds match exactly the log marginal likelihood. 
%\end{remark}

\begin{remark}
The diagonal matrix $\bV^*$ (with the optimal $v_i^*$ values in its diagonal) is the inverse obtained after zeroing out the off-diagonal elements of $\bI + \frac{1}{\sigma^2}(\bK_{\f \f} - \bQ_{\f \f})$, 
i.e., $\bV^* = \text{diag}[\bI + \frac{1}{\sigma^2}(\bK_{\f \f} - \bQ_{\f \f})]^{-1}$ which approximates 
$(\bI + \frac{1}{\sigma^2}(\bK_{\f \f} - \bQ_{\f \f}))^{-1}$ 
in \Cref{eq:exact_cov_pfuy}. 
Also note that in the ordering of positive definite matrices it holds $\bV^* \leq \bI$, from which it follows that $q(\f | \bu)$ has smaller covariance than $p(\f | \bu)$ and more accurately approximates the covariance of $p(\f | \bu, \y)$. 
The latter  as implied by \Cref{eq:exact_cov_pfuy}, 
has also smaller covariance than $p(\f | \bu)$. 
\end{remark}

%Finally, as we discuss in related work our bound is also better than the recent bound on the log determinant by \citet{artemevburt2021cglb}.  

\subsection{Predictions
\label{sec:predictions}
} 

To perform 
predictions we will be using 
the same predictive posterior 
from \Cref{eq:variational_posteriorGP}, i.e., 
$
q(\f_* | \y) =
\int p(\f_* | \bu) q(\bu) d \bu, 
$
where the optimal $q^*(\bu)$ (see \Cref{app:detailsSVGP}) 
is exactly the same as in 
the standard SVGP method. The alternative expression (and strictly speaking more appropriate 
since our variational approximation is $q(\f | \bu) q(\bu)$) is given by 
\begin{equation}
q_{high\_cost}(\f_* | \y) 
\! = \! \! \int p(\f_* | \f, \bu) q(\f |  \bu) q(\bu)  d \f d \bu. 
\end{equation}
But this is expensive since it has cost $\mathcal{O}(N^3)$. The reason is that $\int p(\f_* | \f, \bu) q(\f |  \bu) d \f$ does not simplify anymore since $q(\f |  \bu)$ 
is not the conditional GP, which 
means that $p(\f_* | \f, \bu)$ and $q(\f | \bu)$ are not consistent 
under the GP prior. 
Nevertheless, 
$q(\f_* | \y)$ and $q_{high\_cost}(\f_* | \y)$ have exactly the same mean,  since $q(\f | \bu)$ and $p(\f | \bu)$ have the 
same mean, but the tractable $q$ will give higher variances than  $q_{high\_cost}$. 

\subsection{Stochastic Minibatch Training
\label{sec:stochasticopt}}

The initial SVGP method \cite{titsias2009variational} does the training in a batch mode where all data are used in each optimization step. Stochastic optimization using minibatches was proposed by \citet{hensman2013gaussian}.  
Here, we apply our new approximation to this stochastic method. 

We start from  \Cref{eq:newcollapsedbound_with_vis},
and substitute only the optimal values for each $v_i$
without using the optimal setting for $q(\bu)$. This results in the uncollapsed
bound
\begin{align} 
& \sum_{i=1}^N \biggl\{  \E_{q(\bu)} [\log \mathcal{N}(y_i | \bk_{f_i \bu}
\bK_{\bu \bu }^{-1} \bu, \sigma^2 )]  \nonumber \\
& - \frac{1}{2}  \log\left(1+\frac{k_{ii} - q_{ii}}{\sigma^2} \right)  \biggr\} 
  - \text{KL}[q(\bu) || p(\bu)],
\label{eq:newuncollapsedbound}
\end{align}
where  $\bk_{f_i \bu}$ is the $1 \times M$ vector of all kernel 
values between the training input $\bx_i$ and the inducing inputs $\bZ$, while 
the expectation under $q(\bu)$ in the first line is 
analytic; see \citet{hensman2013gaussian}. %and \Cref{app:detailsNewbounds} for details. 
The above bound is strictly better than 
the previous uncollapsed bound in \citet{hensman2013gaussian},
since $- \frac{1}{2 \sigma^2} (k_{ii} - q_{ii}) \leq -\frac{1}{2} \log\left(1 + \frac{k_{ii} - q_{ii}}{\sigma^2} \right)$. 
Based on the above we can apply stochastic gradient methods to optimize 
$q(\bu)$ and the hyperparameters  by subsampling 
data minibatches to deal with the sum over the $N$ training points, i.e.,\ 
at each iteration we use the stochastic ELBO: 
\begin{align} 
& \frac{N}{|\mathcal{B}|} \sum_{i \in \mathcal{B}} \biggl\{  \E_{q(\bu)} [\log \mathcal{N}(y_i | \bk_{f_i \bu}
\bK_{\bu \bu }^{-1} \bu, \sigma^2 )]  \nonumber \\
& - \frac{1}{2}  \log\left(1+\frac{k_{ii} - q_{ii}}{\sigma^2} \right)  \biggr\} 
  - \text{KL}[q(\bu) || p(\bu)],
\label{eq:newuncollapsedbound_minibatch}
\end{align}
where $\mathcal{B}$ denotes a minibatch. 

The most common parametrization of $q(\bu)$
is $q(\bu) = \mathcal{N}(\bu | \m, \bS)$ where  the mean vector $\m$ and covariance matrix $\bS$ are    
variational parameters.  Another popular 
parametrization, for instance used as the default in GPflow \cite{GPflow17}, is the whitened 
parametrization that we consider in our experiments. %see \Cref{app:whiten} for a review.  
For any choice of
$q(\bu)$,  the new bound is always tighter than its corresponding 
 previous uncollapsed bound and requires minor modifications to existing 
 implementations, i.e., to replace the previous  term $- \frac{1}{2 \sigma^2} (k_{ii} - q_{ii})$  with $-\frac{1}{2} \log\left(1 + \frac{k_{ii} - q_{ii}}{\sigma^2} \right)$.

\subsection{Non-Gaussian Likelihoods
\label{sec:nongaussian}}

Consider a factorized  likelihood $p(\y | \f) = \prod_{i=1}^N p(y_ i | f_i)$ 
where  $p(y_ i | f_i)$ is non-Gaussian, e.g., Bernoulli  for binary outputs  
or Poisson for counts.  
In this non-conjugate setting the sparse 
variational GP approximation imposes the same form for the variational distribution, i.e., $q(\f, \bu) = p(\f | \bu) q(\bu)$ 
where $p(\f | \bu)$ is the 
conditional GP prior. As shown in several works \cite{Chai12,hensman2015scalable,lloyd15,Dezfouli15,Sheth15}, this leads to the bound 
 \begin{equation}
 \sum_{i=1}^N 
\E_{q(f_i)} [\log p(y_i | f_i)]   - \text{KL}[q(\bu) || p(\bu)],
\label{eq:standard_nonconjugate_bound}
 \end{equation} 
 where $q(f_i) = \int p(\f  | \bu) q(\bu ) d \f_{-i} d \bu$ is the marginal  over
 $f_i := f(\bx_i)$  with respect to the approximate posterior $q(\f, \bu)$. Given 
 that  $q(\bu)$ is  Gaussian with mean $\m$ and covariance 
 $\bS$,  $q(f_i)$ can be computed fast in $\mathcal{O}(M^2)$ time (after precomputing the Cholesky factorization of
 $\bK_{\bu \bu}$) as follows 
 \begin{equation}
 q(f_i)  = \mathcal{N}(f_i | \bk_{f_i \bu} \bK_{\bu \bu}^{-1} \m, k_{ii} - q_{ii} + \bk_{f_i \bu} \bK_{\bu \bu}^{-1} \bS \bK_{\bu \bu}^{-1} \bk_{\bu f_i}). 
 \end{equation}
For the discussion next it is useful to observe that the efficiency when computing $q(f_i)$ comes from $p(\f | \bu)$ being a conditional GP prior, so 
expressing $p(f_i | \bu)$ is trivial. 

Suppose now that we wish to impose the more structured variational 
approximation $q(\f, \bu) = q(\f | \bu)  q(\bu)$ where
$q(\bu) = \mathcal{N}(\bu | \m, \bS)$ and $q(\f | \bu)$ 
is given by 
\Cref{eq:qfu}. The bound %(see \Cref{app:nonGaussian}) 
can be written as
\begin{align}
& \sum_{i=1}^N 
\E_{q(f_i)} [\log p(y_i | f_i)] - 
 \frac{1}{2} \sum_{i=1}^N (v_i - \log v_i - 1)
\nonumber \\
& - \text{KL}[q(\bu) || p(\bu)],
\label{eq:nonGaussian_bound_intractable}
\end{align}
where we used the fact that 
$\text{KL}[q(\f|\bu) || p(\f|\bu)]$ is obtained from  \Cref{lem:KLqfupfu}. The above bound
is not computationally efficient since 
the marginal $q(f_i) = \int q(\f  | \bu) q(\bu ) d \f_{-i} d \bu$ 
has $\mathcal{O}(N^3)$ cost. This  is because
the marginalization $q(f_i | \bu) = \int q(\f  | \bu) d \f_{-i}$ cannot be trivially expressed, due to the complex structure of the covariance
$(\bK_{\f \f} - \bQ_{\f \f})^{\frac{1}{2}} \bV
(\bK_{\f \f} - \bQ_{\f \f})^{\frac{1}{2}}$ in $q(\f | \bu)$. To overcome this,  we will use a simplified version of $q(\f | \bu)$, in which we choose a spherical $\bV = v \bI$ with $v > 0$. Then, things become tractable. 

\begin{proposition} Let \emph{$q(\f|\bu) = \mathcal{N}(\f | \bK_{\f \bu} \bK_{\bu \bu}^{-1} \bu, v (\bK_{\f \f} - \bQ_{\f \f}))$} for \emph{$v>0$}. Then (\ref{eq:nonGaussian_bound_intractable}) is computed in \emph{$\mathcal{O}(N M^2)$} time as 
\emph{\begin{align}
& \sum_{i=1}^N 
\E_{q(f_i)} [\log p(y_i | f_i)] -  
 \frac{N}{2} (v - \log v - 1)
\nonumber \\ & - \text{KL}[q(\bu) || p(\bu)],
\label{eq:nonGaussian_bound_tractable}
\end{align}}

\noindent where the marginal is \emph{$q(f_i)  = \mathcal{N}(f_i | \bk_{f_i \bu} \bK_{\bu \bu}^{-1} \m, v (k_{ii} - q_{ii}) + \bk_{f_i \bu} \bK_{\bu \bu}^{-1} \bS \bK_{\bu \bu}^{-1} \bk_{\bu f_i})$}. 
\end{proposition}
%\begin{remark}
The parameter $v$ multiplies the term 
$k_{ii} - q_{ii}$  inside the variance of 
$q(f_i)$, and it also appears in the regularization term
$-\frac{N}{2} (v - \log v - 1)$. If $v=1$ the bound 
in (\ref{eq:nonGaussian_bound_tractable}) reduces to 
(\ref{eq:standard_nonconjugate_bound}), while by
optimizing over $v$ it can become a tighter bound. 
The optimization 
of $v$ is done jointly  
with the remaining parameters $\m,\bS, \bZ, \theta$ using gradient-based methods. Stochastic gradients can also be used 
by subsampling minibatches 
to deal with the sum  
$\sum_{i=1}^N 
\E_{q(f_i)} [\log p(y_i | f_i)]$ and reduce the complexity to $\mathcal{O}(M^3)$. 
%\end{remark}

The above framework can be extended to non-conjugate models having multiple functions, such as  multi-class GP classification, by introducing a separate $v$ parameter per GP function. 
In our experiments, we consider only
single-function non-conjugate GP models and we leave the experimentation with more complex models for future work.

%Finally, note  that we try to use diagonal $\bV$ for the non-Gaussian likelihood case and approximate the marginal $q(f_i)$ by 
%$q(f_i)  = \mathcal{N}(f_i | \bk_{f_i \bu} \bK_{\bu \bu}^{-1} \m, v_i (k_{ii} - q_{ii}) + \bk_{f_i \bu} \bK_{\bu \bu}^{-1} \bS \bK_{\bu \bu}^{-1} \bk_{\bu f_i})$
%(which is not the correct marginal under $q(\f|\bu) q(\bu)$ since the variance term $v_i (k_{ii} - q_{ii})$ is wrong), then this creates an inconsistency in the variational distribution in the ELBO, since the $q(f_i)$ used to compute the expected log-likelihood term will be inconsistent with the $q(\f|\bu) q(\bu)$ in the KL divergence term $KL[q(f|u) q(u) || p(f|u) p(u)]$), and the objective is not a rigorous ELBO anymore.

%% file: sec_related.tex
\section{Related Work
\label{sec:relatedwork}
}

Several recent works on sparse GPs focus on constructing  efficient inducing points, such as works that place inducing points on a grid \citep{wilson2015kernel,evans2018scalable,gardner2018product}, construct inter-domain Fourier features \cite{gredilla09,hensmanetal2018}, provide Bayesian treatments to inducing inputs \cite{rossi21a}
or use nearest neighbor 
sparsity structures
\cite{tran21a, wu22h}.
There exist also algorithms that  allow to increase the number of inducing points using the decoupled method \citep{cheng2017variational, havasi2018deep}
and the related orthogonally decoupled approaches 
\citep{salimbeni2018orthogonally, shietal2020, sun2021, tiao2023}. 
Our contribution is orthogonal to these previous methods since 
we relax the conditional GP
prior
%, $p(\f | \bu)$, 
assumption in the posterior variational approximation. This means that our method could be used to improve previous variational sparse GP approaches,
as the  ones mentioned above as well as earlier schemes that select inducing points from the training inputs \cite{Cao2013,Chai12,Schreiter2016}. 

\citet{XinranZhu2023} proposed 
inducing points GP approximations that change the conditional GP $p(\f | \bu)$ in the variational approximation to a modified conditional GP that uses different kernel hyperparameters in its mean vector. 
Note that our method differs since
our $q(\f | \bu)$ directly tries to 
construct a better approximation to the exact posterior
$p(\f | \bu, \y)$, using the extra $\bV$ variational parameters, 
without changing the kernel hyperparameters; see \Cref{sec:proposedmethod}. More 
importantly, our method has $\mathcal{O}(N M^2)$ cost, while the ELBO in  \citet{XinranZhu2023} (see Section 3.1 and Appendix A.1 in their paper) has cubic cost $\mathcal{O}(N^3)$ since it depends
on the inverse of $\bK_{\f \f} - \bQ_{\f \f}$ (denoted as $\tilde{\bK}_{nn}$ in their paper).  %\citet{XinranZhu2023}). 

\citet{artemevburt2021cglb}
%, by applying linear algebra operations, 
derived an upper bound on the log determinant $\log |\bK_{\f \f} + \sigma^2 \bI|$ in the exact GP log marginal likelihood and obtained the following tighter upper bound to the initial trace 
regularization term $-\frac{1}{2 \sigma^2} \text{tr}\left(  \bK_{\f\f} - \bQ_{\f\f} \right)$: 
\begin{equation}
- \frac{N}{ 2} \log\left( 1 + 
\frac{\text{tr}(\bK_{\f\f} - \bQ_{\f\f})}{N \sigma^2} \right).
\label{eq:artemevbound}
\end{equation}
Our bound is tighter since 
from Jensen's inequality 
it holds $ - \frac{N}{ 2} \log\left( 1 + 
\frac{\text{tr}(\bK_{\f\f} - \bQ_{\f\f})}{N \sigma^2} \right)
\leq - \frac{1}{2} \sum_{i=1}^N \log\left( 1 + 
\frac{k_{ii} - q_{ii}}{\sigma^2} \right)$. Further, the above regularization term
can be interpreted as a restricted special case of our method, obtained through a $q(\f | \bu)$ from \Cref{eq:qfu} where the diagonal matrix $\bV$ is constrained to be spherical $\bV = v  \bI$; see \Cref{app:artemevbound}. Finally note, 
that unlike (\ref{eq:artemevbound}) 
(where the sum is inside the logarithm) our bound allows to apply stochastic optimization
as described in \Cref{sec:stochasticopt}.

Finally, 
\citet{Buietal2017}
used  power expectation 
propagation that minimizes  $\alpha$-divergence and derived an approximation 
to the log marginal likelihood that interpolates between the FITC ($\alpha=1$) log marginal 
likelihood \cite{Snelson2006,candela-rasmussen-05} and the standard collapsed 
variational bound in (\ref{eq:collapsedbound_old})
($\alpha \rightarrow 0$).
This approximation uses the regularization 
term 
\begin{equation}
-\frac{1-\alpha}{2 \alpha}
\sum_{i=1}^N \log\left( 1 + 
\alpha \frac{k_{ii} - q_{ii}}{\sigma^2} \right). 
\label{eq:Buiregularization}
\end{equation}
This is different from 
ours since
there is no value of $\alpha$ 
such that the two regularization terms will become equal. 
For example, note that for $\alpha \rightarrow 0$, \Cref{eq:Buiregularization} 
reduces to $-\frac{1}{2 \sigma^2} \text{tr}\left(  \bK_{\f\f} - \bQ_{\f\f} \right)$ as discussed 
%in Section 3.6 
in \citet{Buietal2017}. 

%% file: sec_experiments.tex
\section{Experiments
\label{sec:experiments}
}

\begin{figure*}[t]
\centering
\begin{tabular}{ccc}
\includegraphics[scale=0.29]
{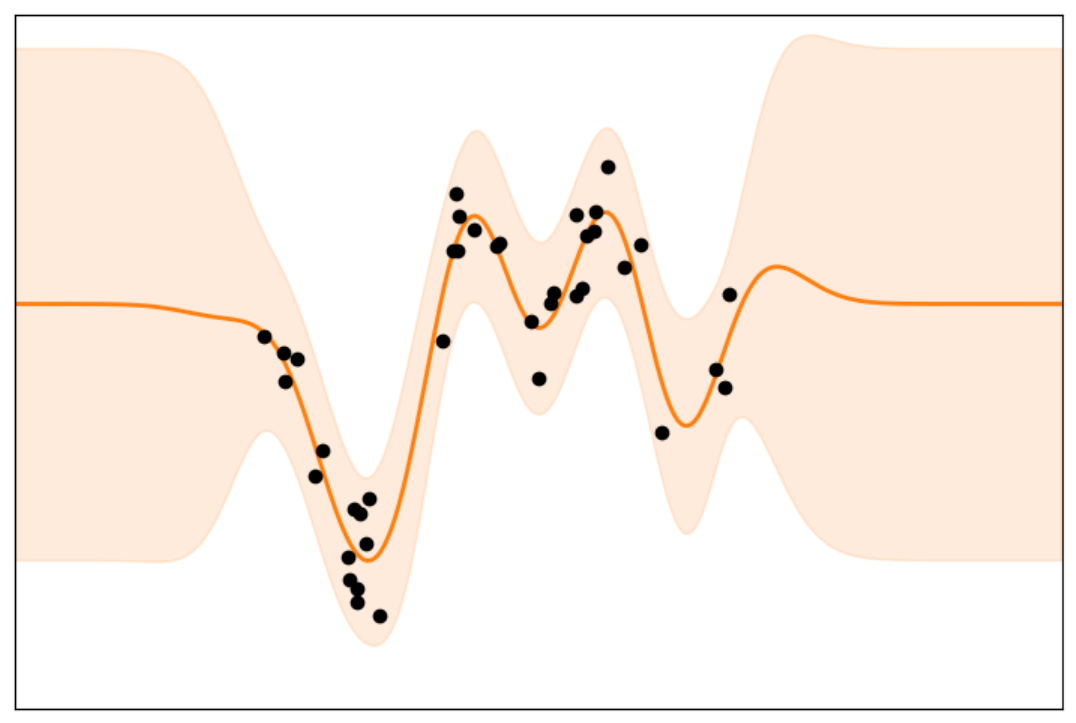} &
\includegraphics[scale=0.29]
{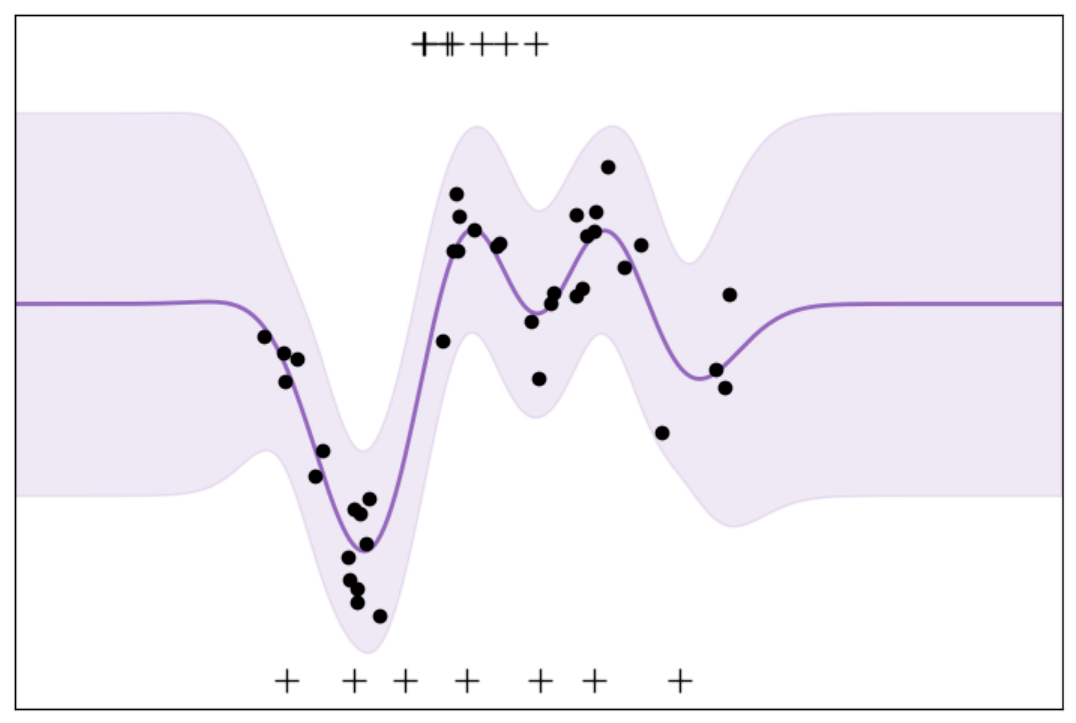} &
\includegraphics[scale=0.29]
{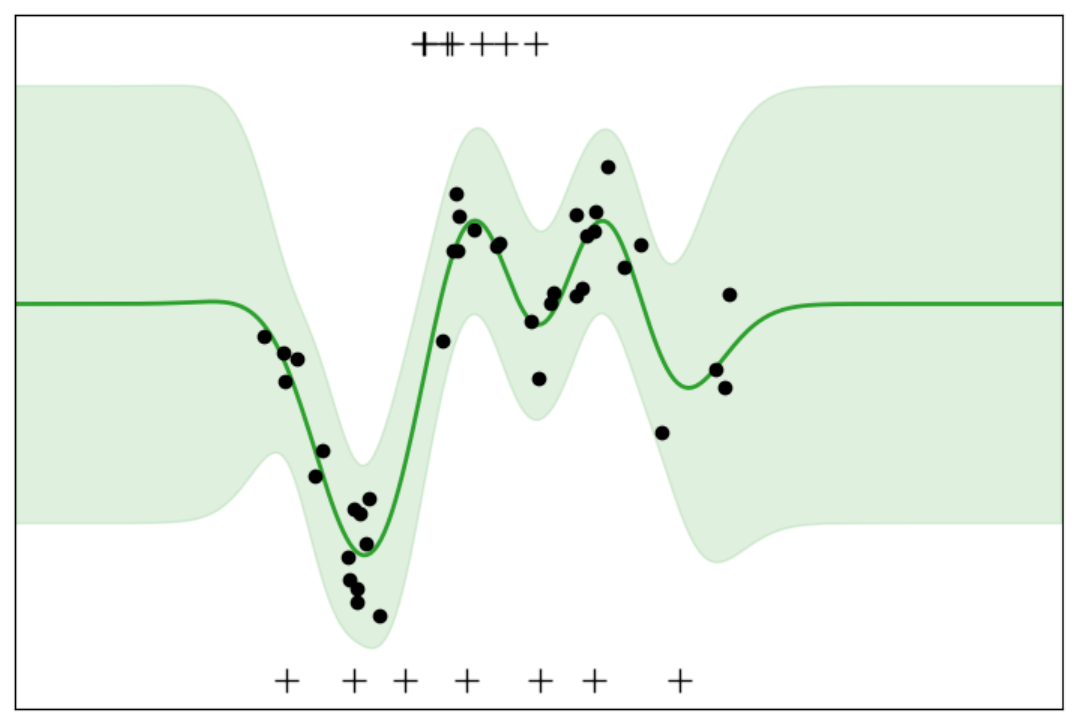} \\
(a) & (b) & (c) \\                
\includegraphics[scale=0.29]
{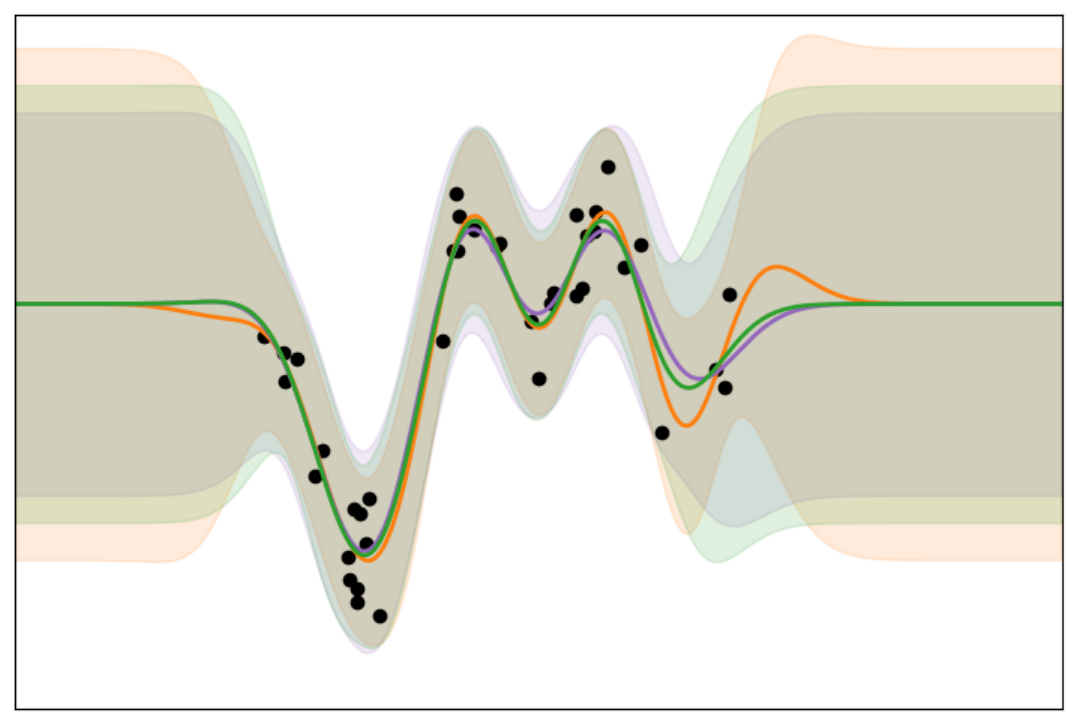} &
\includegraphics[scale=0.25]
{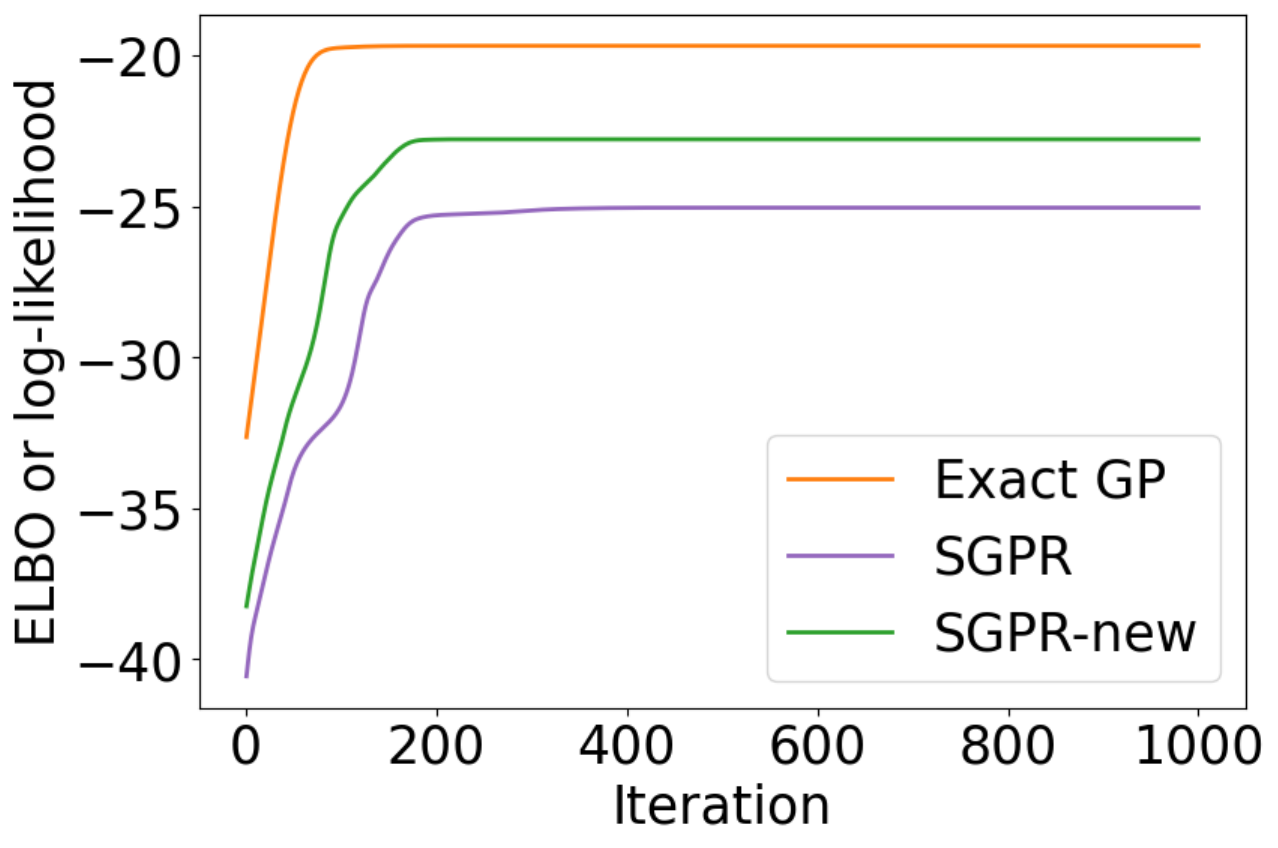} &
\includegraphics[scale=0.25]
{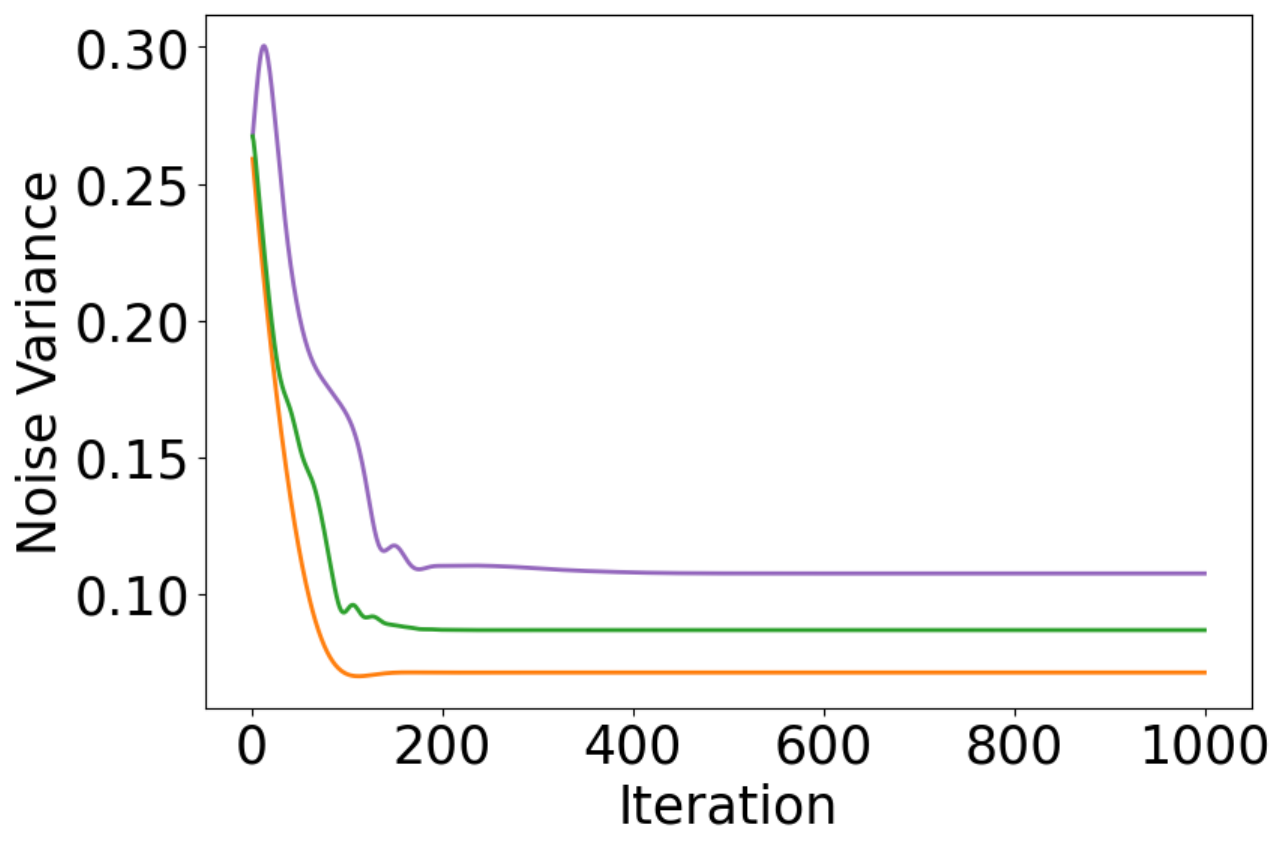} \\
(d) & (e) & (f)              
\end{tabular}
\caption{First row shows posterior predictions (means with 2-standard deviations) after
  fitting the exact GP (a), and the sparse GPs with either the standard collapsed SGPR bound (b) or the proposed SGPR-new collapsed bound (c). In panels (b),(c) the seven inducing points are initialized to the same random locations (shown on top with crosses) while the optimized values are shown at the bottom.
  Panel (d) superimposes all predictions in order to provide a more comparative visualization.
  Finally, panel (e) shows the ELBO (or exact log marginal likelihood for the exact GP) values across optimization steps while (f) shows the corresponding values for the noise variance $\sigma^2$.}
\label{fig:toy}
\end{figure*}

\subsection{Illustration in 1-D Regression}

In the first regression experiment we consider the 1-D  Snelson dataset \cite{Snelson2006}. We took a subset of 40 examples of this dataset and we fitted the exact GP with the squared exponential kernel $k(x, x') = \sigma_f^2 \exp( - \frac{ (x - x')^2}{2 \ell^2})$. We also fitted sparse variational GPs %, denoted as SGPR, 
with either the standard collapsed bound \cite{titsias2009variational} from \Cref{eq:collapsedbound_old} (SGPR) or the new collapsed bound from \Cref{eq:newcollapsedbound} (SGPR-new).
Both sparse GP methods use seven inducing points initialized at the
same values as shown in Figure \ref{fig:toy}. All methods are initialized to the same hyperparameter values; see \Cref{app:furtherresults}.

Figure \ref{fig:toy} shows the results. %Specifically,
Note that both SGPR and SGPR-new find similar inducing point locations. But SGPR-new,  as a tighter bound (see panel (e)), is able to reduce some bias when estimating
the hyperparameters since it finds a noise variance $\sigma^2$ closer to the one by exact GP (see panel (f)).  
This results in better predictions that match better the exact GP, as shown by the comparative visualization in panel (d). From panel (d), observe that both the mean and variances of SGPR-new are closer to the exact GP than SGPR.

\subsection{Medium Size Regression Datasets
\label{sec:mediumregress}
}

\begin{figure*}[t]
\centering
\begin{tabular}{ccc}
\includegraphics[scale=0.25]
{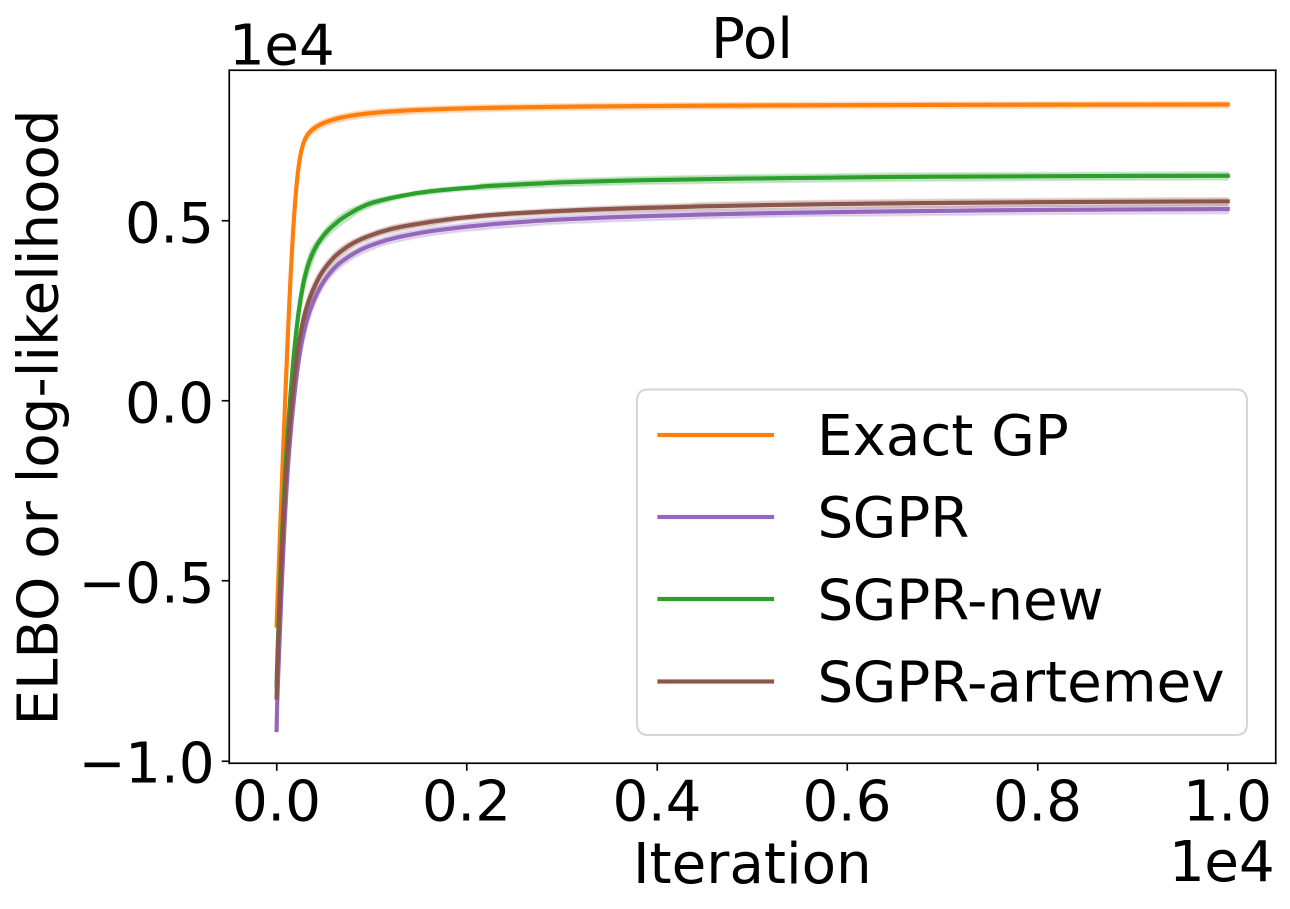} &
\includegraphics[scale=0.25]
{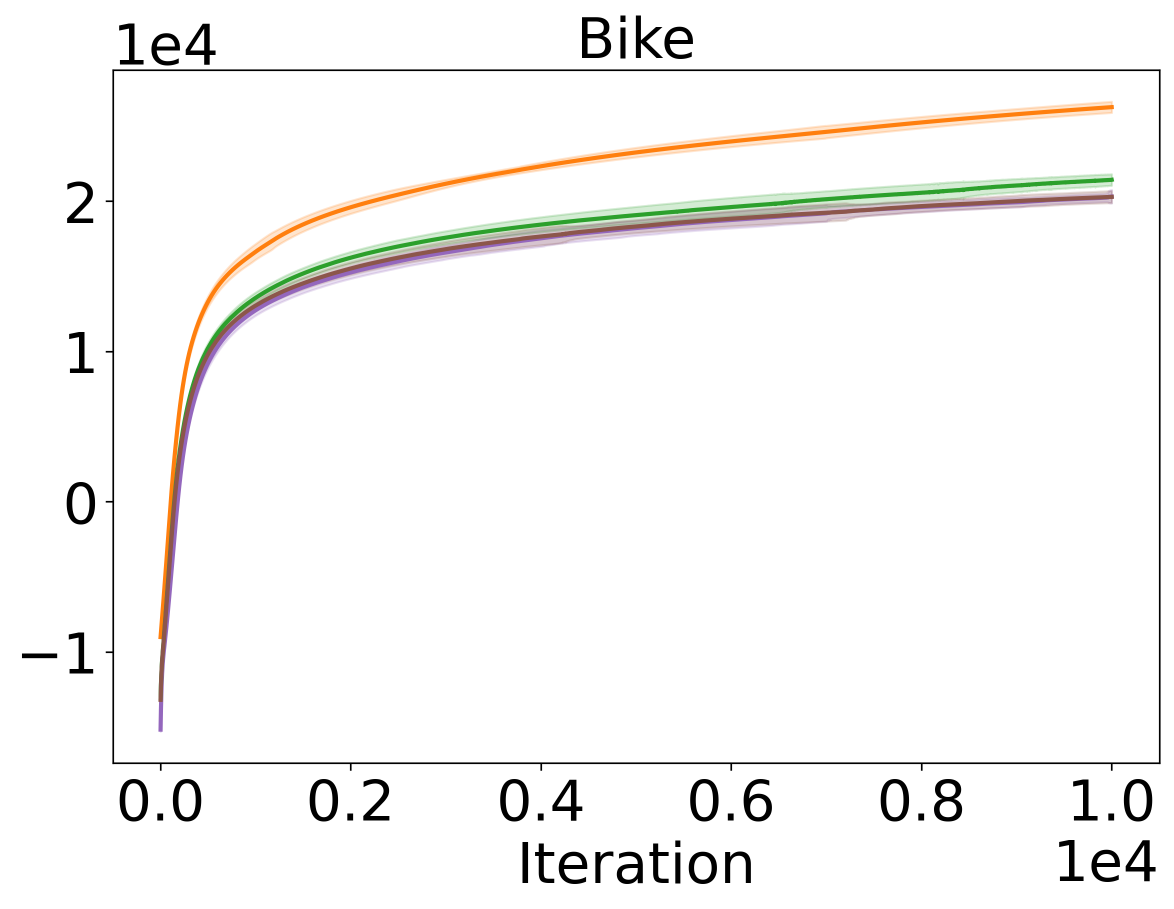} &
\includegraphics[scale=0.25]
{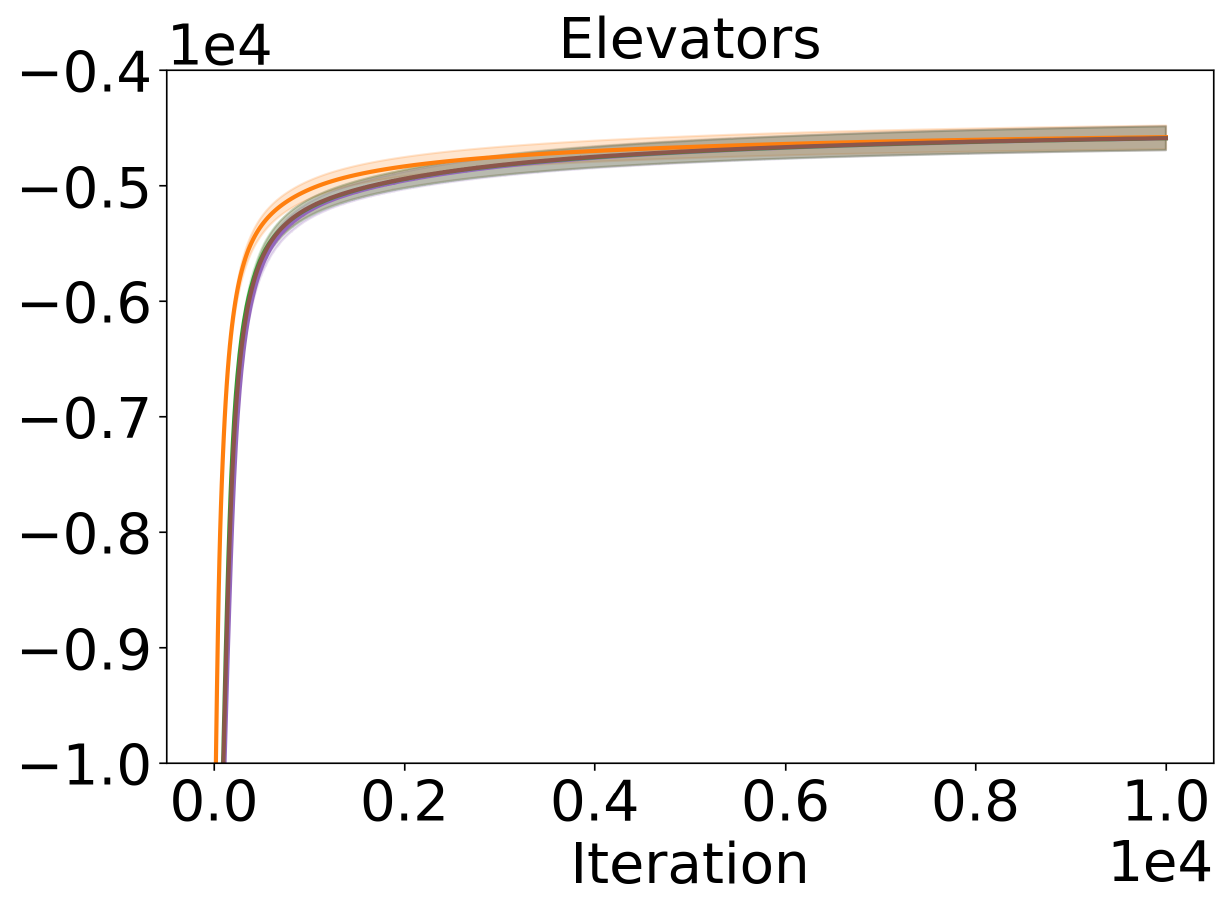} \\
\includegraphics[scale=0.25]
{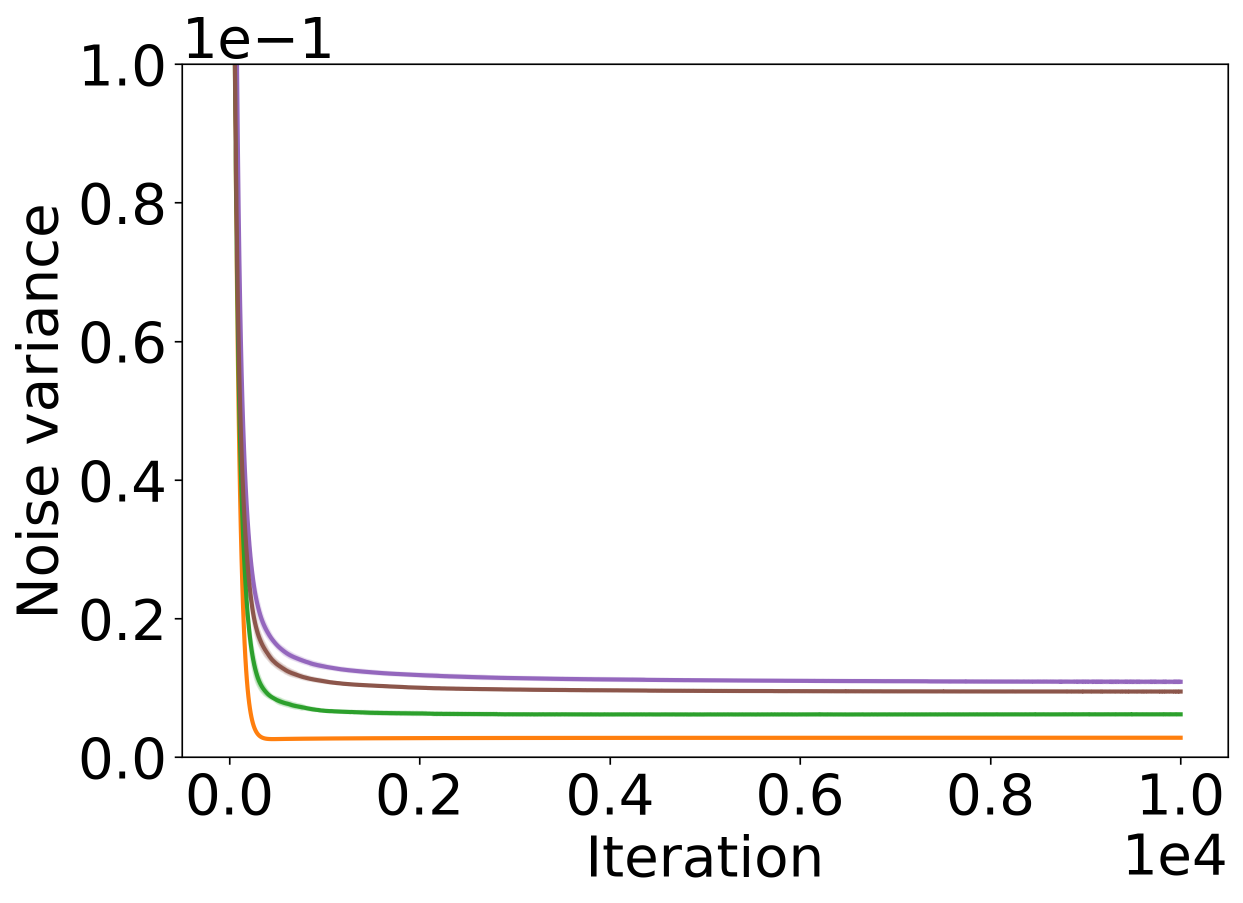} &
\includegraphics[scale=0.25]
{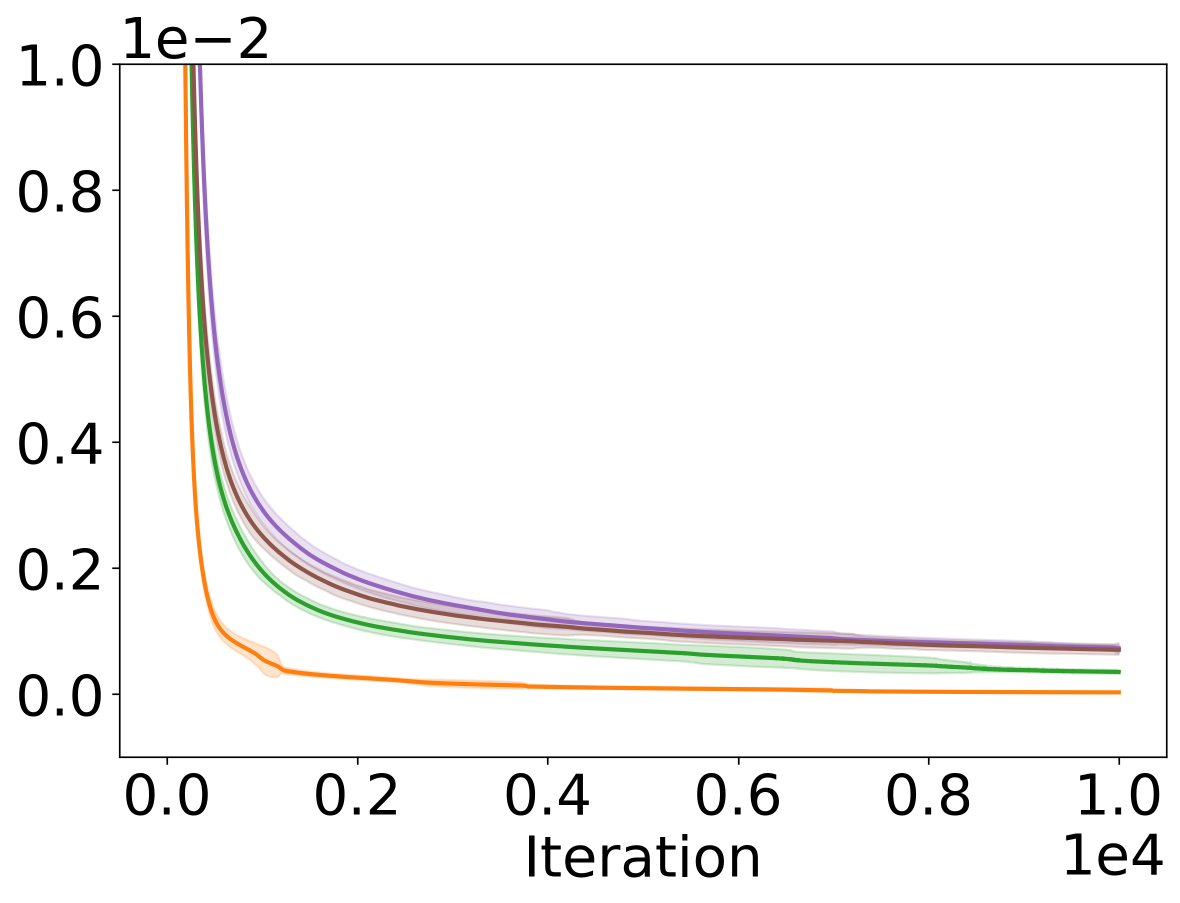} &
\includegraphics[scale=0.25]
{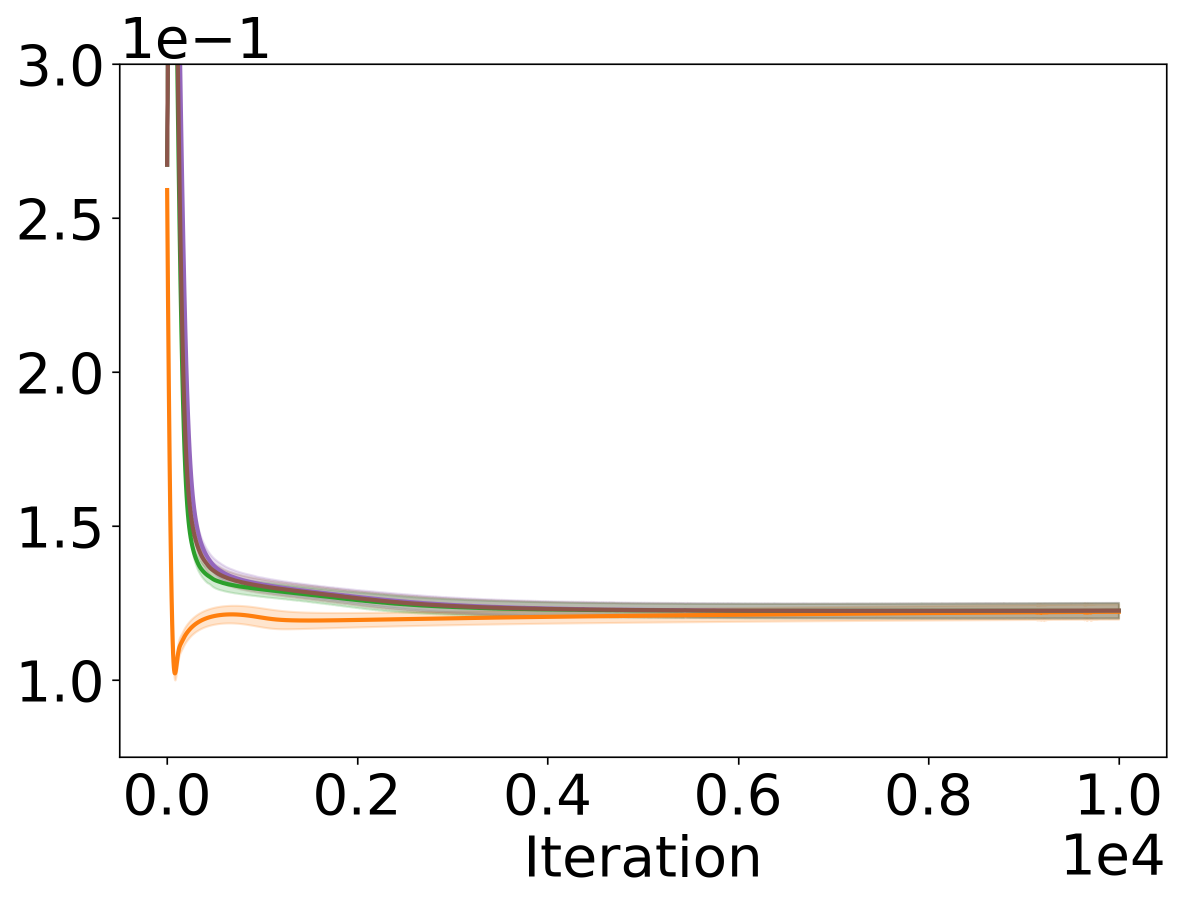} \\
\includegraphics[scale=0.24]
{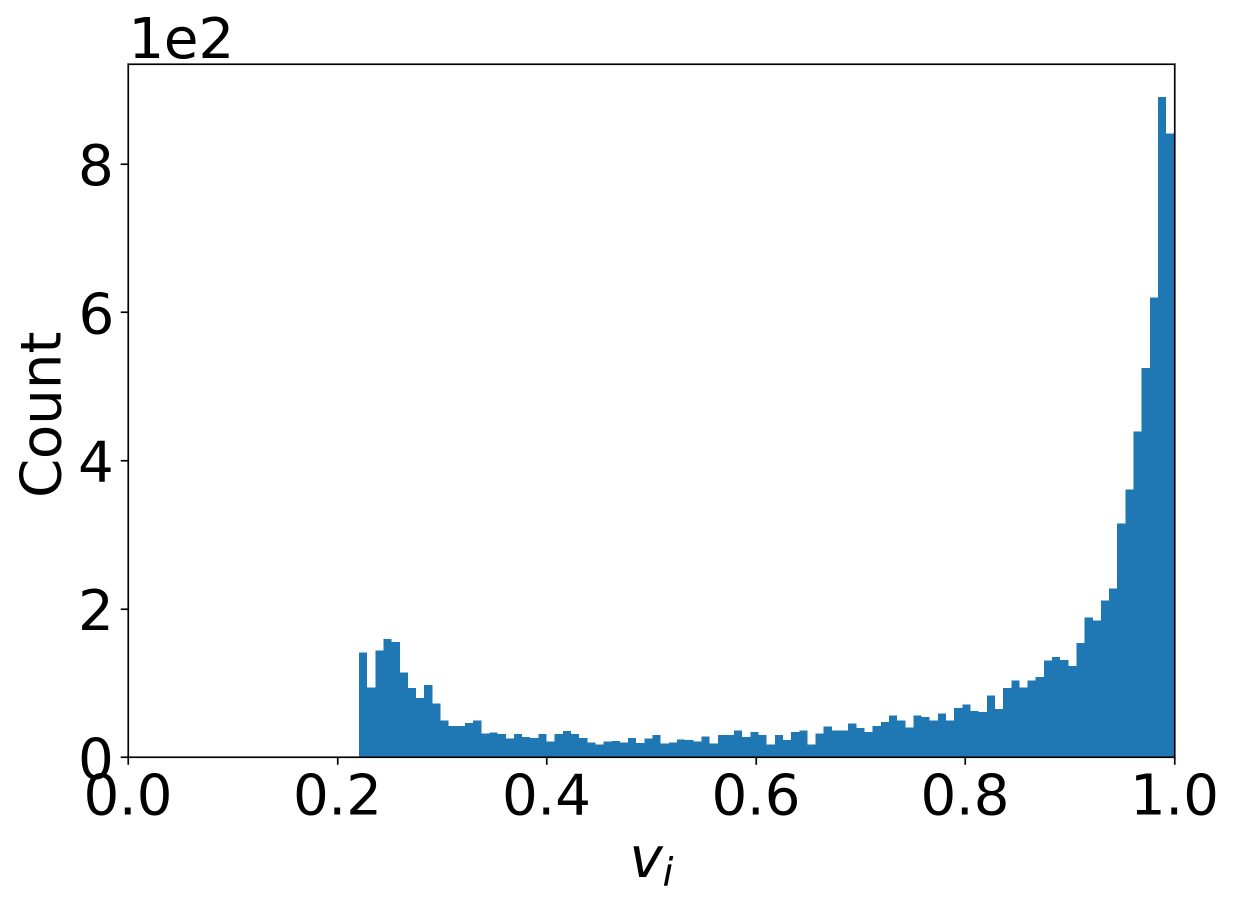} &
\includegraphics[scale=0.24]
{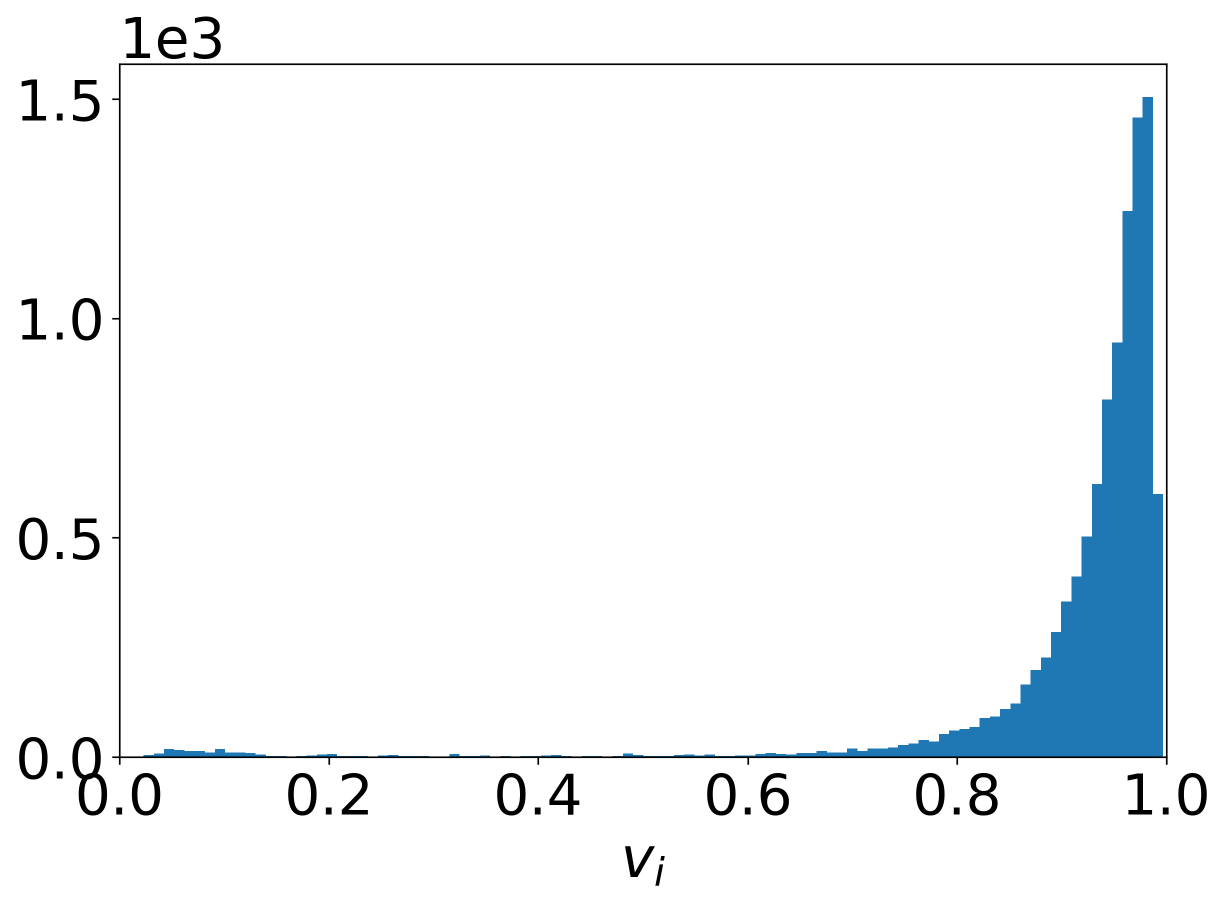} &
\includegraphics[scale=0.24]
{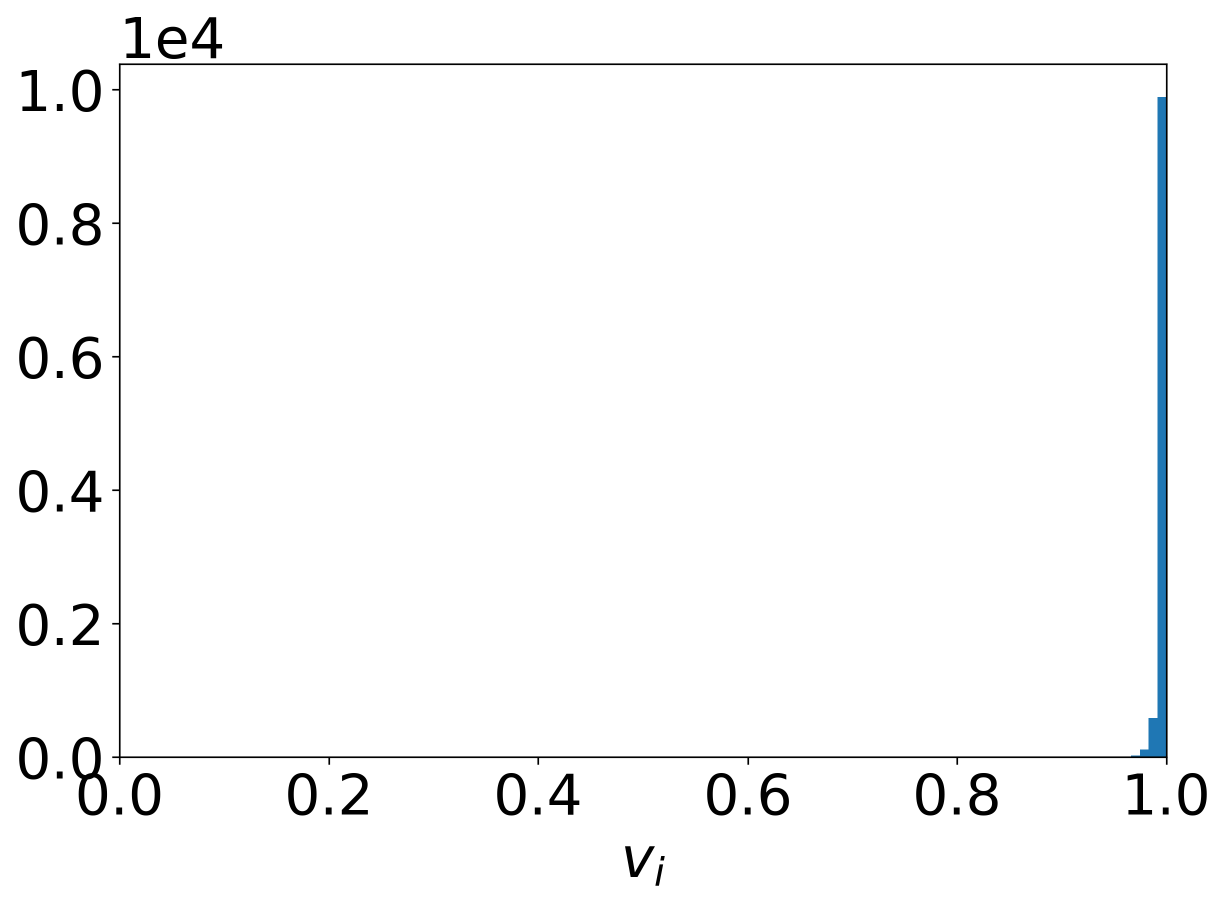}
\end{tabular}
\caption{The three plots in each column correspond to the same dataset: first row shows the ELBO (or log-likelihood)
 for all four methods (Exact GP, SGPR, SGPR-new and SGPR-artemev) with the number of iterations, and the plot in the second row shows the
  corresponding values for $\sigma^2$. SGPR methods use $M=1024$ inducing points initialized by k-means. For these two first lines we plot the mean and standard error
  after repeating the experiment five times with different train-test dataset splits; see \Cref{app:furtherresults} for further experimental details. For one of the runs of SGPR-new, the third line shows histograms for  the estimated values of the variational parameters $v_i = \left(1 + \frac{k_{ii} - q_{ii}}{\sigma^2} \right)^{-1}$.
  }
\label{fig:mediumsize1024}
\end{figure*}

To further investigate the findings from the previous section, we consider three medium size real-world UCI regression datasets (Pol, Bike, and Elevators)
with roughly 10k training data points each, and for which we can still run the exact GP. We choose the ARD squared exponential kernel $k(\bx, \bx') = \sigma_f^2 \exp( - \sum_{i=1}^d \frac{(x_i - x_i')^2}{2 \ell_i^2})$.
We run all three previous methods (Exact GP, SGPR, SGPR-new) five times with different random train-test splits;
see \Cref{app:furtherresults} for experimental details. We also include
in the comparison a fourth method (discussed in Related Work)
which is the \citet{artemevburt2021cglb}'s bound  (SGPR-artemev) that does training using the collapsed bound from \Cref{eq:artemvecollapsedbound} in \Cref{app:artemevbound}. 
All sparse GP methods use $M=1024$ or $M=2048$ inducing points initialized by k-means.  Figure \ref{fig:mediumsize1024} (in the first two lines) shows the objective function and the noise variance $\sigma^2$ across $10k$ optimization steps using Adam with base learning rate $0.01$ and for $M=1024$.
For SGPR-new, the third line in  \Cref{fig:mediumsize1024}
shows histograms of the estimated final values of the optimal $v_i$ variational parameters. 
\Cref{fig:mediumsize2048} in \Cref{app:mediumsizeRegress} shows the corresponding plots for $M=2048$.  We observe that for Pol and Bike, SGPR-new matches closer the exact GP training than SGPR and SGPR-artemev. Specifically, SGPR-new gives higher ELBO and estimates the noise variance with reduced underfitting bias.
For the Elevators dataset, $M=1024$ inducing points were enough for sparse GP methods to closely match exact GP training. This happens because in this case $\bQ_{\f \f}$ accurately approximates $\bK_{\f \f}$, i.e., the elements $k_{ii} - q_{ii}$ get close to zero. For this latter dataset, observe that since the $k_{ii} - q_{ii}$
values are close to zero 
the corresponding $v_i$ values are concentrated around one as shown by the corresponding (right-most) 
histogram in \Cref{fig:mediumsize1024}.

Table \ref{table:smalldatasetsTestLL} reports test log-likelihood predictions which show that 
SGPR-new outperforms SGPR and SGPR-artemev.  

\begin{table}
  \caption{Average test log-likelihoods for the medium size regression datasets.
  The numbers in parentheses are standard errors.
    %The SGPR methods used $M=1024$ inducing points.
  }
\label{table:smalldatasetsTestLL}
%\vskip 0.15in
%\begin{small}
\begin{center}
%  \begin{sc}
\resizebox{\linewidth}{!}{%
\begin{tabular}{lcccr}
\toprule
& Pol  & Bike & Elevators \\
\midrule
Exact GP & $1.089(0.011)$ & $3.105(0.022)$ & $-0.386(0.001)$ \\
% Exact GP & $1.089(0.011)$ & $3.105(0.022)$ & $-0.386(0.001)$  \\
\midrule
 $M=1024$ & & & \\
SGPR & $0.821(0.008)$ & $2.176(0.020)$ & $-0.387(0.001)$\\
% SGPR-trace & $0.958(0.008)$  & $2.337(0.030)$ & $-0.387(0.001)$ \\
SGPR-artemev & $0.859(0.007)$ & $2.199(0.024)$ & $-0.387(0.001)$  \\
SGPR-new & $0.920(0.006)$ & $2.326(0.026)$  & $-0.387(0.001)$  \\
%SGPR-log & $0.998(0.008)$  & $2.511(0.021)$ & $-0.387(0.001)$ \\
\midrule
$M=2048$ & & & \\
% SGPR-trace & $0.821(0.008)$ & $2.176(0.020)$ & $-0.387(0.001)$\\
SGPR & $0.958(0.008)$  & $2.337(0.030)$ & $-0.387(0.001)$ \\
% SGPR-log & $0.920(0.006)$ & $2.326(0.026)$  & $-0.387(0.001)$  \\
SGPR-artemev & $0.976(0.008)$ & $2.356(0.029)$ & $-0.387(0.001)$  \\
SGPR-new & $0.998(0.008)$  & $2.511(0.021)$ & $-0.387(0.001)$ \\
\bottomrule
\end{tabular}}
%\end{sc}
%\end{small}
\end{center}
%\vskip -0.1in
\end{table}

\subsection{Large Scale Regression Datasets
\label{sec:largeregress}
}

\begin{table*}[t]
\caption{Test log-likelihoods for the large scale regression datasets with standard errors in parentheses. Best mean values are highlighted.} 
% Uses random 80\% / 20\% training and test splits, repeated 5 times. }
\label{table:largescaleTestLL}
\makebox[\textwidth][c]{
\resizebox{1.02\textwidth}{!}{
\setlength\tabcolsep{2pt}
\begin{tabular}{ l l cc cc cc cc}
\toprule
& & Kin40k &  Protein & \footnotesize KeggDirected & KEGGU &  3dRoad & Song &  Buzz & \footnotesize HouseElectric \\
\cmidrule(lr){3-10}
& $N$ & 25,600 & 29,267 & 31,248 & 40,708 & 278,319 & 329,820 & 373,280 & 1,311,539  \\
& $d$ & 8 & 9 & 20 & 27 & 3 & 90 & 77 & 9  \\
\midrule
%\multirow{2}{*}{SVGP}
%& $1024$  
%& 0.094(0.003) & -0.963(0.006) & 0.967(0.005) & 0.678(0.004) & -0.698(0.002) & -1.193(0.001) & -0.079(0.002) & 1.304(0.002)  \\
%& $1536$  
%& 0.129(0.003) & -0.949(0.005) & 0.944(0.006) & 0.673(0.004) & -0.674(0.003) & -1.193(0.001) & -0.079(0.002) & 1.304(0.003) \\
%\midrule
From \citet{shietal2020} \\ 
ODVGP & $1024+1024$ 
& 0.137(0.003) & -0.956(0.005) & -0.199(0.067) & 0.105(0.033) & -0.664(0.003) & -1.193(0.001) & -0.078(0.001) & 1.317(0.002) \\
& $1024+8096$  
& 0.144(0.002) & -0.946(0.005) & -0.136(0.063) & 0.109(0.033) & -0.657(0.003) & -1.193(0.001) & -0.079(0.001) & 1.319(0.004) \\
SOLVE-GP & $1024 + 1024$ & 0.187(0.002) & -0.943(0.005) &  0.973(0.003) &  0.680(0.003) & -0.659(0.002) & -1.192(0.001) &  -0.071(0.001) & 1.333(0.003) \\
%\midrule
%SVGP
% \\
%& $2048$
%& 0.137(0.003) & {\bf -0.940}(0.005) & 0.907(0.003) & 0.665(0.004) & -0.669(0.002) & {\bf -1.192}(0.001) & -0.079(0.002) & 1.304(0.003) \\
\midrule
SVGP & 1024 & $0.108(0.002)$ & $-0.969(0.006)$ & $1.042(0.009)$ & $0.699(0.005)$ & $-0.704(0.003)$ & $-1.192(0.001)$ & $-0.069(0.002)$ & $1.383(0.002)$ \\
& 2048 & $0.237(0.002)$ & $-0.944(0.006)$ & ${\bf 1.050}(0.009)$ & ${\bf 0.703}(0.005)$ & ${\bf -0.650}(0.003)$ & ${\bf -1.190}(0.001)$ & $-0.063(0.001)$ & $1.419(0.002)$ \\
SVGP-new  & 1024 & $0.152(0.003)$ & $-0.965(0.006)$ & $1.044(0.009)$ & $0.699(0.005)$ & $-0.701(0.003)$ & $-1.192(0.001)$ & $-0.065(0.002)$ & $1.387(0.003)$ \\
 & 2048 & ${\bf 0.286}(0.002)$ & ${\bf -0.938}(0.006)$ & $1.051(0.009)$ & ${\bf 0.703}(0.005)$ & $-0.651(0.004)$ & ${\bf -1.190}(0.001)$ & ${\bf -0.060}(0.001)$ & ${\bf 1.421}(0.002)$ \\
\bottomrule 
\end{tabular}
}
}
\end{table*}

We consider 8 regression datasets, with training data sizes ranging from tens of thousands to millions. 
%Results of exact GP regression have been reported on these datasets with distributed training~\citep{wang2019exact}. 
We implemented the stochastic optimization versions of the two scalable sparse GP methods: (i) the one that trains using the previous uncollapsed bound from
 \citet{hensman2013gaussian} (SVGP) and (ii) our new bound from    
\Cref{eq:newuncollapsedbound} (SVGP-new). We denote these stochastic optimization versions by SVGP to distinguish them from the corresponding
SGPR methods that use the more expensive collapsed bounds. We run the SVGP methods with $M=1024$ and $2048$ inducing points, Matern3/2 kernel with common lengthscale, minibatch size $1024$, Adam with base learning rate $0.01$ and $100$ epochs. These experimental settings match the ones in \citet{wang2019exact} and \citet{shietal2020} as further described  in \Cref{app:largescaleRegress}. Table \ref{table:largescaleTestLL} reports the test log likelihood scores
for all datasets. In the comparison we also included two strong baselines from Table 2 in \citet{shietal2020}, i.e., SOLVE-GP and ODVGP \cite{salimbeni2018orthogonally}.

\begin{figure*}
\centering
\begin{tabular}{ccc}
\includegraphics[scale=0.24]
{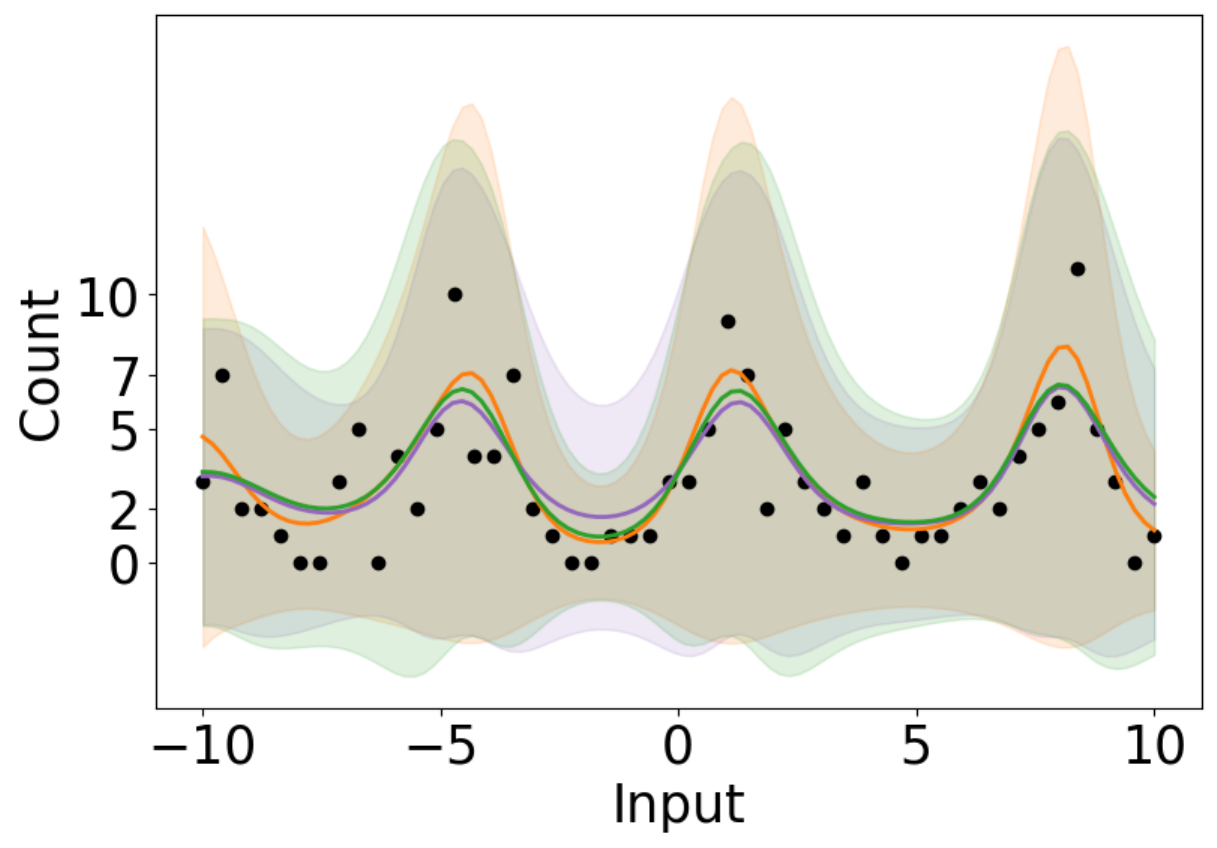} &
\includegraphics[scale=0.24]
{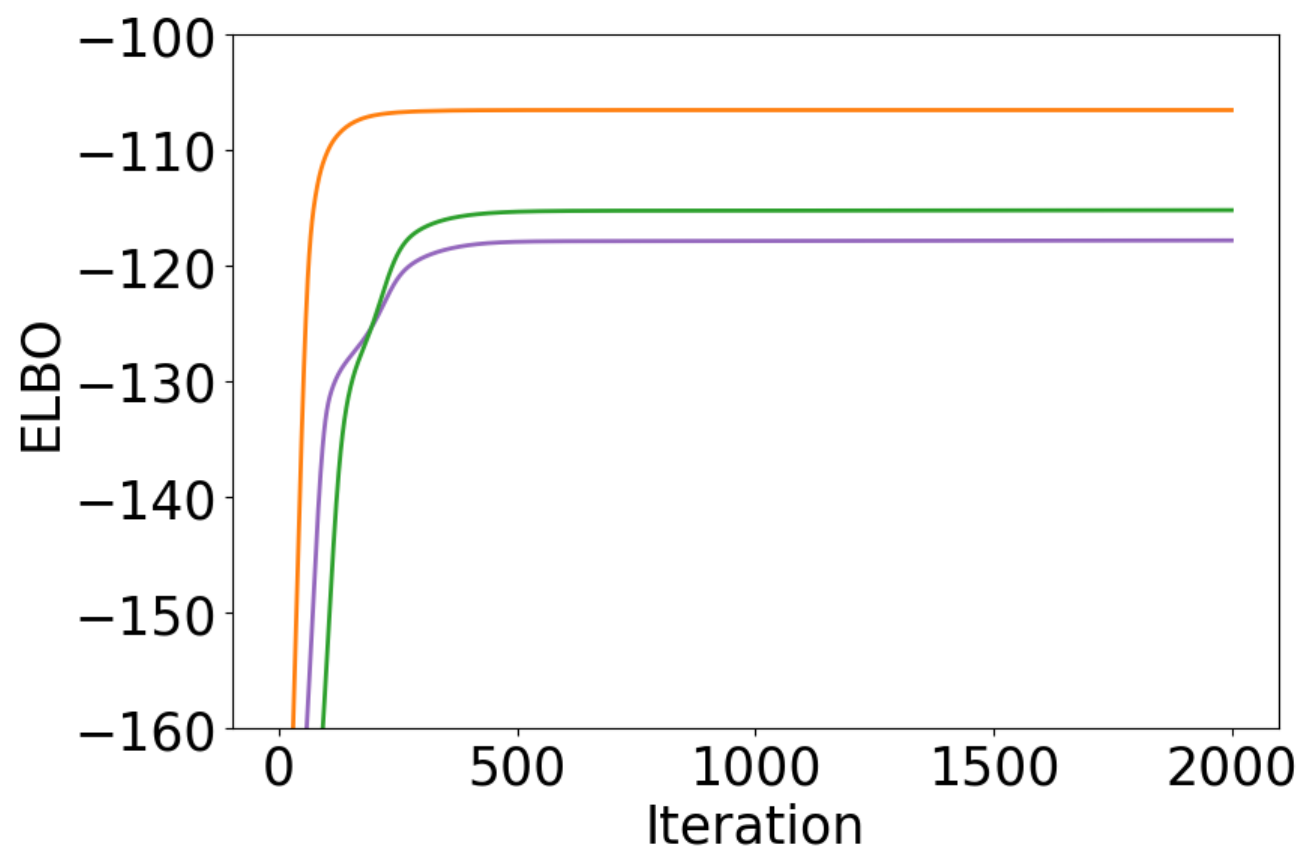} &
\includegraphics[scale=0.24]
{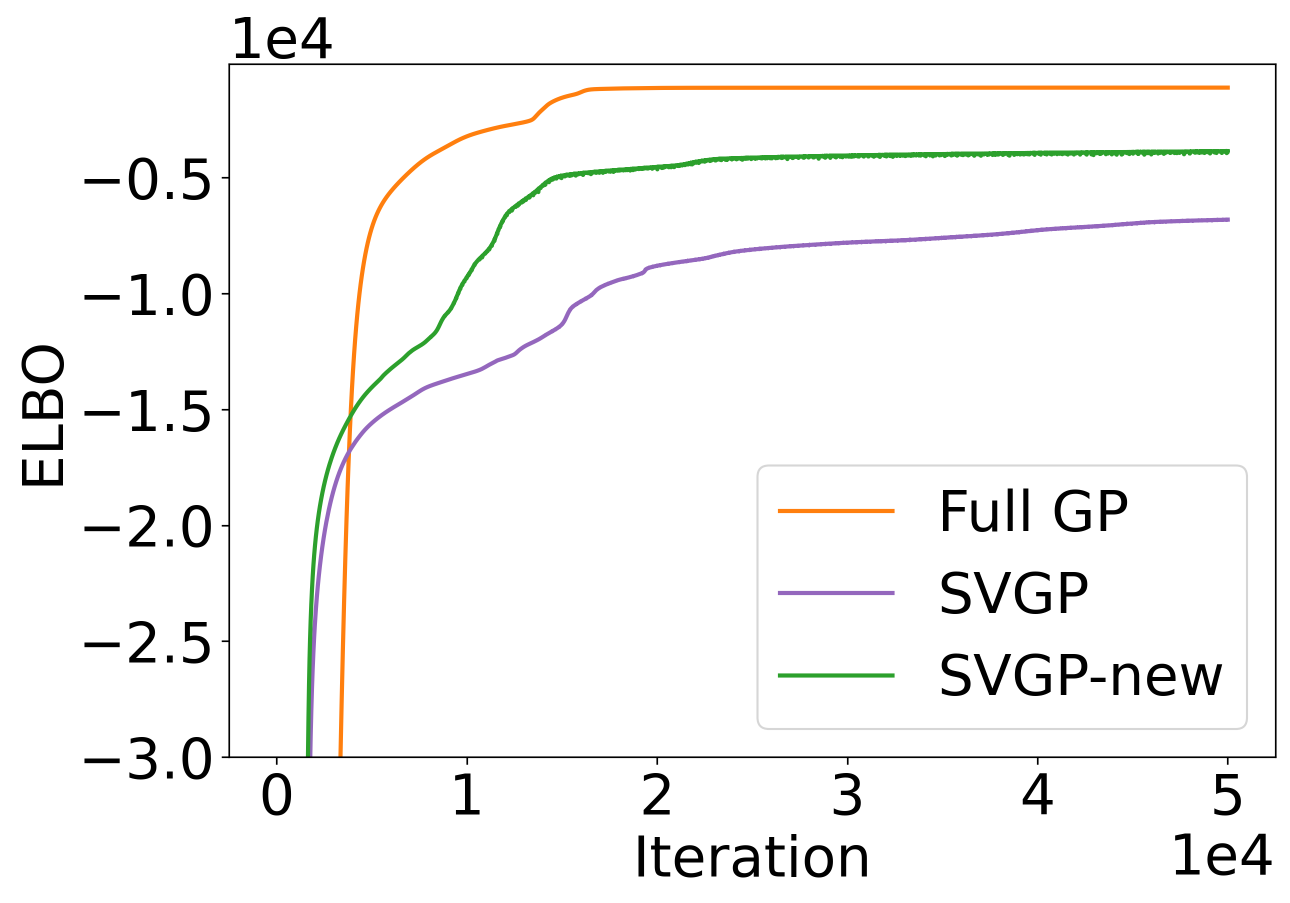} \\
% (a) & (b) & (c)
\end{tabular}
\caption{({\bf left}) shows the % posterior 
predictions (means with 2-standard deviations) over counts (black dots) in the artificial data example  after
  fitting the Full GP, and the two SVGPs. This plot superimposes all predictions in order to provide a comparative visualization.
  %; see \Cref{app:poisson} for individual plots. 
  ({\bf middle})  shows the ELBO  across optimization steps for the artificial data example. ({\bf right}) shows the ELBO for the NYBikes dataset and $M=16$.}
\label{fig:poisson}
\end{figure*}

From the predictive log likelihood scores in Table \ref{table:largescaleTestLL} and also the corresponding Root Mean Squared Error (RMSE)
scores reported in  \Cref{table:largescaleRMSE} in \Cref{app:largescaleRegress}, we can conclude that training with the new SVGP-new variational bound
provides a clear improvement compared to training with the previous SVGP bound. Note that this improvement requires no change in the computational
cost, and in fact there is only a minor modification needed to be done in an existing SVGP implementation in order to run SVGP-new.  

\subsection{Poisson Regression
  \label{sec:poisson}
}

We consider a non-Gaussian likelihood example where the output data are counts modeled  by a Poisson likelihood 
$p(\y | \f) = \prod_{i=1}^N \frac{e^{f_i}}{y_i !} e^{-e^{f_i}}$  where the log intensities values follow a GP prior. For such 
case the new variational approximation includes a single additional variational parameter denoted by $v$, which is optimized together 
with the remaining parameters; see \cref{sec:nongaussian}. We will compare training with the new ELBO 
 from \Cref{eq:nonGaussian_bound_tractable}  (we denote this method by SVGP-new) with the standard ELBO that is obtained by restricting  $v=1$  (SVGP). 
 
Firstly, we consider an artificial example of $50$ observations with 1-D inputs placed in the grid $[-10, 10]$ where counts are
generated using Poisson intensities given by $\lambda(x) = 3.5 + 3  \sin(x)$. We train the GP model with the SVGP bound and the proposed SVGP-new bound using $6$ inducing points initialized to the same values for both methods; see \Cref{app:poisson}. 
\Cref{fig:poisson}(left) shows the 
 observed counts together with the 
predictions obtained by SVGP, SVGP-new 
and non-sparse %or full 
variational % inference 
GP (Full GP). From
this figure and from the
ELBO values, 
we observe that SVGP-new
remains closer to Full GP.  

Secondly, we consider a real dataset (NYBikes) about bicycles crossings going over bridges in New York City\footnote{This dataset is freely available from
\url{https://www.kaggle.com/datasets/new-york-city/nyc-east-river-bicycle-crossings}.}.
This dataset is a daily record of the number of bicycles crossing into or out of Manhattan via one of the East River bridges over a period 9 months. The data contains $210$  points and we randomly choose $90\%$ for training and $10\%$ for test.   
We apply GP Poisson regression for the Brooklyn bridge counts where the input vector $\bx$ is taken to be two-dimensional consisted of 
 maximum and minimum daily temperatures.  We train the sparse GPs with either SVGP or SVGP-new and with $M=8,16,32$ 
 inducing points initialized by k-means.  Since the dataset is small  we also run the non-sparse  Full GP. The ELBO across iterations in \Cref{fig:poisson} (right) and the test log likelihood scores (\Cref{table:poisson_nybikes} 
 in \Cref{app:poisson})
 indicate that  SVGP-new provides a better approximation than SVGP.

%% file: sec_conclusion.tex
\section{Conclusions
\label{sec:conclusions}
}

We have presented a method that relaxes the conditional GP assumption in the approximate 
distribution in sparse variational GPs.
This leads to tighter collapsed and uncollapsed bounds, that maintain the computational cost with the previous bounds and can reduce training underfitting. For future work 
an interesting topic is to apply our method to more complex GP models, 
such as those with multiple outputs, with uncertain inputs and deep GPs. 
For the Bayesian GP-LVM, where the collapsed closed form bound has strong
similarities with the previous GP regression collapsed bound in \Cref{eq:collapsedbound_old}, deriving  
a new collapsed bound is tractable as described in \Cref{app:bgplvm}. 
Finally, it might be useful to investigate whether theoretical convergence results on sparse GPs
\cite{burt2020convergence,wild2023etal}, can be improved given the new collapsed lower bound.

%% file: appendix.tex
\section{Further details about the SVGP method
\label{app:detailsSVGP}
}

We give a brief overview of the derivation of the standard collapsed bound 
in \Cref{eq:collapsedbound_old}. 
Some steps of the derivation will also be instructive for proving the 
main results of this paper in 
\Cref{app:detailsNewbounds}.  

Given the variational distribution
$q(\f,\bu) = p(\f| \bu) q(\bu)$  the lower bound is   
\begin{align} 
\log p(\y) & \geq 
\int p(\f|\bu) q(\bu)
\log \frac{p(\y | \f) \cancel{p(\f|\bu)}  p(\bu)}
{\cancel{p(\f|\bu)} q(\bu)} d \f d \bu \nonumber \\ 
& = 
\int p(\f|\bu) q(\bu)
\log \frac{p(\y | \f) p(\bu)}
{q(\bu)} d \f d\bu \nonumber  \\
& =  
\int q(\bu) \left\{
 \int p(\f|\bu)
\log p(\y |\f) d \f + 
\log \frac{p(\bu)} {q(\bu)} \right\} d \bu \nonumber \\ 
& = \int q(\bu) \log \frac{\exp\{ \int p(\f | \bu) \log p(\y | \f) d \f\} p(\bu)}{q(\bu)} d \bu.
\label{eq:appendixbound1}
\end{align}
The expectation $\int p(\f|\bu)
\log p(\y |\f) d \f$ can be computed 
as
\begin{align}
\int p(\f|\bu)
\log p(\y |\f) d \f & = 
\int p(\f|\bu) 
 \log p(\y | \f) d \f \nonumber \\ 
& = \int p(\f|\bu)\left\{
- \frac{N}{2}\log(2 \pi \sigma^2) 
- \frac{1}{2 \sigma^2} \text{tr}\left[ 
\y \y^T 
- 2 \y \f^T 
+ \f \f^T 
\right] \right\} d \f \nonumber \\
& =  
- \frac{N}{2}\log(2 \pi \sigma^2) 
- \frac{1}{2 \sigma^2} \text{tr}\left[ 
\y \y^T 
- 2 \y (\bK_{\f \bu} \bK_{\bu \bu}^{-1}\bu)^\top 
+ (\bK_{\f \bu} \bK_{\bu \bu}^{-1}\bu) (\bK_{\f \bu} \bK_{\bu \bu}^{-1}\bu)^\top + \textcolor{blue}{\bK_{\f \f} - \bQ_{\f \f}} \right] \nonumber \\  
& = \log \left[ 
\mathcal{N}(\y|\bK_{\f \bu} \bK_{\bu \bu}^{-1}\bu,\sigma^2 \bI) \right]  
- \frac{1}{2\sigma^2} \text{tr}(\bK_{\f \f} - \bQ_{\f \f}). 
\label{eq:appendixpfulogpyf}
\end{align}
where we highlighted with blue 
a term in the third line to contrast it 
with a similar term when proving \Cref{lem:Expqfu_loglik} in \Cref{app:artemevbound}. The ELBO in \Cref{eq:appendixbound1} 
is written as
\begin{align}
 \log p(\y) & \geq 
\int q(\bu)
\log \frac{ \mathcal{N}(\y | \bK_{\f \bu} \bK_{\bu \bu}^{-1}\bu ,\sigma^2 \bI) p(\bu)}
{q(\bu)} d \f  - \frac{1}{2 \sigma^2} \text{tr}(\bK_{\f \f} - \bQ_{\f \f}).
\label{eq:appendixBound2}
\end{align} 
By maximizing this bound wrt the distribution $q(\bu)$ we obtain the
optimal $q^*$:
\begin{align}
\label{eq:appendixOptimalqu1}
q^*(\bu) & = \frac{ \mathcal{N}(\y |\bK_{\f \bu} \bK_{\bu \bu}^{-1}\bu,\sigma^2
  \bI) p(\bu)} {\int \mathcal{N}(\y | \bK_{\f \bu} \bK_{\bu \bu}^{-1}\bu,\sigma^2
  \bI) p(\bu) d \bu} 
  = \frac{ \mathcal{N}(\y |\bK_{\f \bu} \bK_{\bu \bu}^{-1}\bu,\sigma^2
  \bI) p(\bu)} {\mathcal{N}(\y | {\bf 0}, \bQ_{\f \f}  + \sigma^2 
  \bI)} \\ 
 & =  
 \mathcal{N}(\bu | \sigma^{-2} \bK_{\bu \bu} (\bK_{\bu \bu} + \sigma^{-2} \bK_{\bu \f} \bK_{\f \bu})^{-1} \bK_{\bu \f} \y, 
\bK_{\bu \bu} (\bK_{\bu \bu} + \sigma^{-2} \bK_{\bu \f} \bK_{\f \bu})^{-1} \bK_{\bu \bu}), 
\end{align}
 where the expression in the second line (obtained after some standard 
 completion of a square procedure) shows that $q^*(\bu)$ can be computed in $\mathcal{O}(N M^2)$ time. In fact, this optimal $q^*(\bu)$
 is the same as the one obtained by the DTC (also known as projected process) approximation \cite{seeger-williams-lawrence-03a,candela-rasmussen-05}. By substituting 
 the expression in (\ref{eq:appendixOptimalqu1}) into the bound in (\ref{eq:appendixBound2})
 we obtain the well-known formula of the collapsed bound: 
 \begin{align}
 \log p(\y) & \geq 
\log \mathcal{N}(\y |{\bf 0}, \bQ_{\f \f}  + \sigma^2 \bI)   - \frac{1}{2 \sigma^2} \text{tr}(\bK_{\f \f} - \bQ_{\f \f}).
\label{eq:appendixBound3}
\end{align} 
Given the Gaussian
form of $q(\bu) = \mathcal{N}(\bu|\bfmu, \bA)$ the posterior GP  
is given by $q(\f_*) = \int p(\f_* | \bu) q(\bu) d \bu$:
\begin{align}
q(\f_*) = \mathcal{N}(\f_* | 
\bK_{\f_* \f} \bK_{\bu \bu}^{-1}\bfmu,  \bK_{\f_* \f_*} 
- \bK_{\f_* \bu} \bK_{\bu \bu}^{-1} \bK_{\bu \f_*}  +   \bK_{\f_* \bu} \bK_{\bu \bu}^{-1} \bA \bK_{\bu \bu}^{-1}  \bK_{\bu \f_*} )
\end{align}
which further simplifies if we substitute the optimal mean and covariance of $q^*(\bu)$:
\begin{align}
q(\f_*) = \mathcal{N}(\f_* | 
\bK_{\f_* \f} \bLambda^{-1} \bK_{\bu \f} \frac{\y}{\sigma^2},  \bK_{\f_* \f_*} 
- \bK_{\f_* \bu} \bK_{\bu \bu}^{-1} \bK_{\bu \f_*}  +   \bK_{\f_* \bu}  \bLambda^{-1} \bK_{\bu \f_*} )
\end{align}
where $\bLambda = \bK_{\bu \bu} + \sigma^{-2} \bK_{\bu \f} \bK_{\f \bu}$.

\section{Further details about the proposed bounds
\label{app:detailsNewbounds}
}

Here, we provide several proofs
regarding the proposed bounds.

\subsection{Proof of \Cref{lem:KLqfupfu}} 

$q(\f | \bu)$ 
and $p(\f | \bu)$ are Gaussian distributions having the same mean but different covariance matrices. Thus, 
the KL divergence reduces to 
\begin{align}
\text{KL}[q(\f | \bu) || p(\f | \bu)] 
& = \frac{1}{2} \left\{ 
\log \frac{|\bK_{\f \f} - \bQ_{\f \f}|}{|(\bK_{\f \f} - \bQ_{\f \f})^{\frac{1}{2}} \bV (\bK_{\f \f} - \bQ_{\f \f})^{\frac{1}{2}}|}  
- N + 
\tr\{ (\bK_{\f \f} - \bQ_{\f \f})^{-1}  (\bK_{\f \f} - \bQ_{\f \f})^{\frac{1}{2}} \bV  (\bK_{\f \f} - \bQ_{\f \f})^{\frac{1}{2}} \}  \right\} \nonumber \\ 
& = \frac{1}{2} \left\{ 
- \log | \bV|  
- N + 
\tr\{\bV  \}  \right\},
\end{align}
where all the terms involving 
$\bK_{\f \f} - \bQ_{\f \f}$ cancel out by using standard properties of the 
matrix determinant and trace. Now since $\bV$ is a diagonal matrix (with diagonal elements $v_i>0$) we
conclude that 
$\text{KL}[q(\f | \bu) || p(\f | \bu)] 
= \frac{1}{2} \sum_{i=1}^N (v_i - \log v_i - 1)$. 

\subsection{Proof of \Cref{lem:Expqfu_loglik}}

The derivation of $\int q(\f | \bu) \log p(\y | \f) d \f$ is 
similar to the derivation in \Cref{eq:appendixpfulogpyf} 
with a small difference highlighted  in blue:
\begin{align}
\int p(\f|\bu)
& \log p(\y |\f) d \f \nonumber \\  
& = - \frac{N}{2}\log(2 \pi \sigma^2) 
- \frac{1}{2 \sigma^2} \text{tr}\left[ 
\y \y^T 
- 2 \y (\bK_{\f \bu} \bK_{\bu \bu}^{-1}\bu)^\top 
+ (\bK_{\f \bu} \bK_{\bu \bu}^{-1}\bu) (\bK_{\f \bu} \bK_{\bu \bu}^{-1}\bu)^\top + \textcolor{blue}{(\bK_{\f \f} - \bQ_{\f \f})^{\frac{1}{2}} \bV (\bK_{\f \f} - \bQ_{\f \f})^{\frac{1}{2}}}   \right] \nonumber \\  
& = \log \left[ 
\mathcal{N}(\y|\bK_{\f \bu} \bK_{\bu \bu}^{-1}\bu,\sigma^2 \bI) \right]  
- \frac{1}{2\sigma^2} \text{tr}(\bV (\bK_{\f \f} - \bQ_{\f \f})), 
\label{eq:appendixqfulogpyf}
\end{align}
where we used that $\text{tr}((\bK_{\f \f} - \bQ_{\f \f})^{\frac{1}{2}} \bV (\bK_{\f \f} - \bQ_{\f \f})^{\frac{1}{2}}) = \text{tr}(\bV (\bK_{\f \f} - \bQ_{\f \f}))
$. Now since $\bV$ is a diagonal matrix 
we have $\text{tr}(\bV (\bK_{\f \f} - \bQ_{\f \f})) = \sum_{i=1}^N v_i (k_{ii} - q_{ii})$ which completes the proof. 

\subsection{Proof of \Cref{prop:newbound}
}

The ELBO is written as 
$$ 
 \log p(\y) \geq \int q(\f | \bu) q(\bu) \log \frac{p(\y | \f) p(\f | \bu) p(\bu)}{q(\f | \bu) q(\bu)} =  \int \! \!  q(\bu) \! \left\{ \! \log \frac{\exp\{\E_{q(\f | \bu)}[\log p(\y | \f)]\} p(\bu)}{q(\bu)} \! - \! \text{KL}[q(\f | \bu) || p(\f | \bu)] 
\! \right\} \! d \bu 
$$
and by using the results from 
the two lemmas this becomes 
$$ 
 \log p(\y) \geq \int  q(\bu) \log \frac{  \mathcal{N}(\y | \bK_{\f \bu}
\bK_{\bu \bu }^{-1} \bu, \sigma^2 \bI) p(\bu)}{q(\bu)}  d \bu  - \frac{1}{2} 
\sum_{i=1}^N \left\{  v_i \left(1 + \frac{k_{ii} - q_{ii}}{\sigma^2}\right) - \log v_i -1 \right\}.  
$$
Clearly, maximizing over 
$q(\bu)$ gives the same optimal distribution as in \Cref{eq:appendixOptimalqu1}, and the first term in the bound is the DTC log likelihood. The second term that depends on the $v_i$s is a concave function over these parameters. Thus, by differentiating and setting to zero we obtain the optimal values $v_i^* = \left(1 + \frac{k_{ii} - q_{ii}}{\sigma^2}\right)^{-1}$. If we plug these values back into the bound we obtain the new tighter collapsed bound in \Cref{prop:newbound}.

\subsection{Reinterpretation of \citet{artemevburt2021cglb}'s bound
\label{app:artemevbound}
}

We consider the following form  of $q(\f | \bu)$: 
$$
q(\f|\bu) = \mathcal{N}(\f | \bK_{\f \bu} \bK_{\bu \bu}^{-1} \bu, v (\bK_{\f \f} - \bQ_{\f \f})).
$$
Then, $\text{KL}[q(\f | \bu) || p(\f | \bu)] 
= \frac{N}{2} (v - \log v - 1)$
and $ \E_{q(\f | \bu)}[\log p(\y | \f)] \nonumber \\ 
 = \log \mathcal{N}(\y | \bK_{\f \bu}
\bK_{\bu \bu }^{-1} \bu, \sigma^2 \bI)
- \frac{v}{2 \sigma^2}  \text{tr}(\bK_{\f \f} - \bQ_{\f \f})$ and the bound is written as 
$$ 
 \log p(\y)  \geq \int \! \!  q(\bu) \log \frac{  \mathcal{N}(\y | \bK_{\f \bu}
\bK_{\bu \bu }^{-1} \bu, \sigma^2 \bI) p(\bu)}{q(\bu)}  d \bu - \frac{1}{2}  \left\{  \frac{v}{\sigma^2} \text{tr}(\bK_{\f \f} - \bQ_{\f \f}) +  N (v  -  \log v -1) \right\}.
$$
By maximizing wrt $v$ we obtain $v^* = \left( 1 + \frac{\text{tr}(\bK_{\f \f} - \bQ_{\f \f})}{N \sigma^2} \right)^{-1}$, and by substituting this back into the bound we obtain  \citet{artemevburt2021cglb}'s tighter bound on the initial trace regularization term. Overall this collapsed bound has the form
\begin{equation} 
\log p(\y) \geq \log  \mathcal{N}(\y |{\bf 0},   \bQ_{\f \f} + \sigma^2 \bI) 
 - \frac{N}{2}  \log \left(\ 1 + \frac{\text{tr}(\bK_{\f \f} - \bQ_{\f \f})}{N \sigma^2}  \right).   
\label{eq:artemvecollapsedbound}
\end{equation}
This collapsed bound is what the method
SGPR-artemev is using in \Cref{sec:mediumregress}. Note that 
\citet{artemevburt2021cglb} propose also additional but more expensive bounds for the first DTC log likelihood term that require running conjugate gradients. 
We do not consider those in our comparisons (such bounds could be used in all SGPR bounds since all share the same DTC log likelihood term) as they have higher cost.  

%\section{Non-Gaussian %likelihoods\label{app:nonGaussian}
%}

\subsection{Bound for the Bayesian GP-LVM
\label{app:bgplvm}
}

Due to the strong similarity of the 
standard collapsed SVGP bound in \Cref{eq:collapsedbound_old} with the collapsed 
bound in the Bayesian 
GP-LVM \cite{titsias10a}, applying the new approximation to Bayesian GP-LVM 
seems to be simple and we discuss it next.  

Given observed data $Y \in \Real^{N \times D}$
and latent variables 
$X \in \Real^{N \times Q}$ we have the latent variable model 
$$
p(Y|X) p(X) 
= \left(\prod_{d=1}^D p(\y_d | X) \right) p(X),
$$
where $p(X)$ is a Gaussian prior over the latent variables and 
$p(\y_d | X) 
= \mathcal{N}(\y_d | \bK_{\f \f}(X) + \sigma^2 \bI)$. 
Given a Gaussian variational distribution 
$q(X)$ over the latent variables, the initial form of the bound is 
\begin{align}
F & = \int q(X) \log p(Y|X) d X - \text{KL}[q(X) || p(X)] 
\nonumber \\
& = \sum_{d=1}^D \int q(X) \log p(\y_d|X) d X - \text{KL}[q(X) || p(X)] \nonumber \\
& = \sum_{d=1}^D F_d  - \text{KL}[q(X) || p(X)], 
\end{align}
where $F_d = \int q(X) \log p(\y_d|X) d X$. The KL part will be a tractable KL between two Gaussians, and thus the difficulty is to approximate $F_d$. 
Given that 
$\log p(\y_d | X)$ has the same form with the log marginal likelihood in GP regression, we can lower bound it using inducing variables and exactly the same form of $q(\f|\bu)$ as we did in the main paper. This gives
$$
\log p(\y_d | X) 
\geq  \int q(\bu) \log \frac{  \mathcal{N}(\y_d | \bK_{\f \bu}(X)
\bK_{\bu \bu }^{-1} \bu, \sigma^2 \bI) p(\bu)}{q(\bu)}  d \bu  - \frac{1}{2} 
\sum_{i=1}^N \left\{  v_i \left(1 + \frac{k(\bx_i, \bx_i) - q(\bx_i,\bx_i)}{\sigma^2}\right) - \log v_i -1 \right\},  
$$
where $\bx_i$ is the latent variable for the $i$-th data point and  $q(\bx_i, \bx_i) = \bk(x_i)^\top \bK_{\bu \bu}^{-1} \bk(\bx_i) = \text{tr}\{\bK_{\bu \bu}^{-1} \bk(\bx_i) \bk(\bx_i)^\top\}$. Note that we write the cross  
kernel matrix as 
$\bK_{\f \bu}(X)$ to emphasize its dependence on the latent variables $X$, while $\bK_{\bu \bu}$ does not depend on $X$. Note also that we assume that each $v_i$ parameter does not depend on $\bx_i$ and this is crucial to obtain a closed form collapsed bound. Now we follow 
the derivation in the initial Bayesian GP-LVM
where we use the 
above bound to replace 
$\log p(\y_d|X)$ in 
 $\int q(X) \log p(\y_d|X) d X$ and do first the expectation 
 over $X$, and then solve for the optimal $q(\bu)$. This eliminates $q(\bu)$ and it gives the bound 
 $$
 \log \int e^{\langle \mathcal{N}(\y_d | \bK_{\f \bu}(X)
\bK_{\bu \bu }^{-1} \bu, \sigma^2 \bI)\rangle_{q(X)}} p(\bu) d \bu 
 - \frac{1}{2} 
\sum_{i=1}^N \left\{  v_i \left(1 + \frac{\langle k(\bx_i, \bx_i)\rangle_{q(\bx_i)} -  \langle q(\bx_i,\bx_i) \rangle_{q(\bx_i)}}{\sigma^2}\right) - \log v_i -1 \right\}. 
$$
where we  have used physics notation for expectation, i.e.,  
$\langle \cdot \rangle$. 
For $v_i=1$ this is the previous collapsed  bound used by Bayesian GP-LVM. By maximizing over each $v_i$ we obtain the new collapsed bound 
$$
F_d \geq \log \int e^{\langle \mathcal{N}(\y_d | \bK_{\f \bu}(X)
\bK_{\bu \bu }^{-1} \bu, \sigma^2 \bI)\rangle_{q(X)}} p(\bu) d \bu 
 - \frac{1}{2} 
\sum_{i=1}^N \log \left(1 + \frac{\langle k(\bx_i, \bx_i)\rangle_{q(\bx_i)} -  \langle q(\bx_i,\bx_i) \rangle_{q(\bx_i)}}{\sigma^2}\right), 
$$
which can be substituted in the overall Bayesian GP-LVM bound above. Again the implementation of the new bound requires a minor modification to existing 
Bayesian GP-LVM code.

\section{Learned hyperparameters in 1-D Snelson dataset}

Table \ref{table:hypers_toy} provides the learned hyperparameters for the 1-D Snelson dataset.

\begin{table}[htbp] %\vskip \baselineskip
\caption{Hyperparameter values in 1-D Snelson example. 
} 
\label{table:hypers_toy}
\centering
 \begin{tabular}{llll}
\toprule
& $\sigma^2$ & $\sigma_f^2$  & $\ell^2$ \\
\midrule
Exact GP & 0.0715 & 0.712 & 0.597 \\
SVGP-new & 0.087 & 0.485 & 0.615 \\
SVGP & 0.108 & 0.331 & 0.617 \\
\bottomrule 
\end{tabular}
\vskip \baselineskip
\end{table}

\section{Further experimental details and results
\label{app:furtherresults}
} 

For all regression experiments 
(apart from the toy Snelson 1-D dataset) 
we repeat the runs for five times using different random training and test splits. 
By following \citet{wang2019exact} and \citet{shietal2020} we consider 
80\% / 20\% training / test splits. 
A 20\% subset of the training set is used for validation. 

The training inputs and regression outputs are normalized to have zero mean. For the hyperparameters $\sigma^2, \sigma_f^2, \ell^2$ (or $\ell_i^2$ for ARD kernels) we use the softplus activation to parametrize the square roots of these parameters, i.e., 
to parametrize $\sigma, \sigma_f, \ell_i$. For all experiments we use 
the initializations $\sigma = 0.51$, 
$\sigma_f = 0.69$, $\ell_i=1.0$.  
The inducing inputs $\bZ$ are initialized by running at maximum 30 iterations of k-means clustering with the centers initialized at a random training data subset.  

%All experiments were performed in a V100 GPU.  

\subsection{Medium size regression datasets
\label{app:mediumsizeRegress}
}

For all three datasets in 
this section we can run Exact
GP given the medium training size.
For the Pol dataset the training size 
is $N=9600$ and input dimensionality  
$d=26$. For Elevators is $N=10623$ 
and $d=18$. For the Bike dataset
the initial train size (see e.g., 
Table 7 in \citet{shietal2020}) 
is $N = 11122$ (with $d = 17$) 
but since Exact GP training gave out-of-memory error when running in a V100 GPU, we had to slightly reduce the training size to $N=10600$.  

An mentioned in the main paper the standard squared exponential ARD kernel 
was used in all experiments in this 
section. For training, we perform $10000$ optimization iterations using the Adam optimizer with base learning $0.01$.

\begin{figure*}[t]
\centering
\begin{tabular}{ccc}
\includegraphics[scale=0.25]
{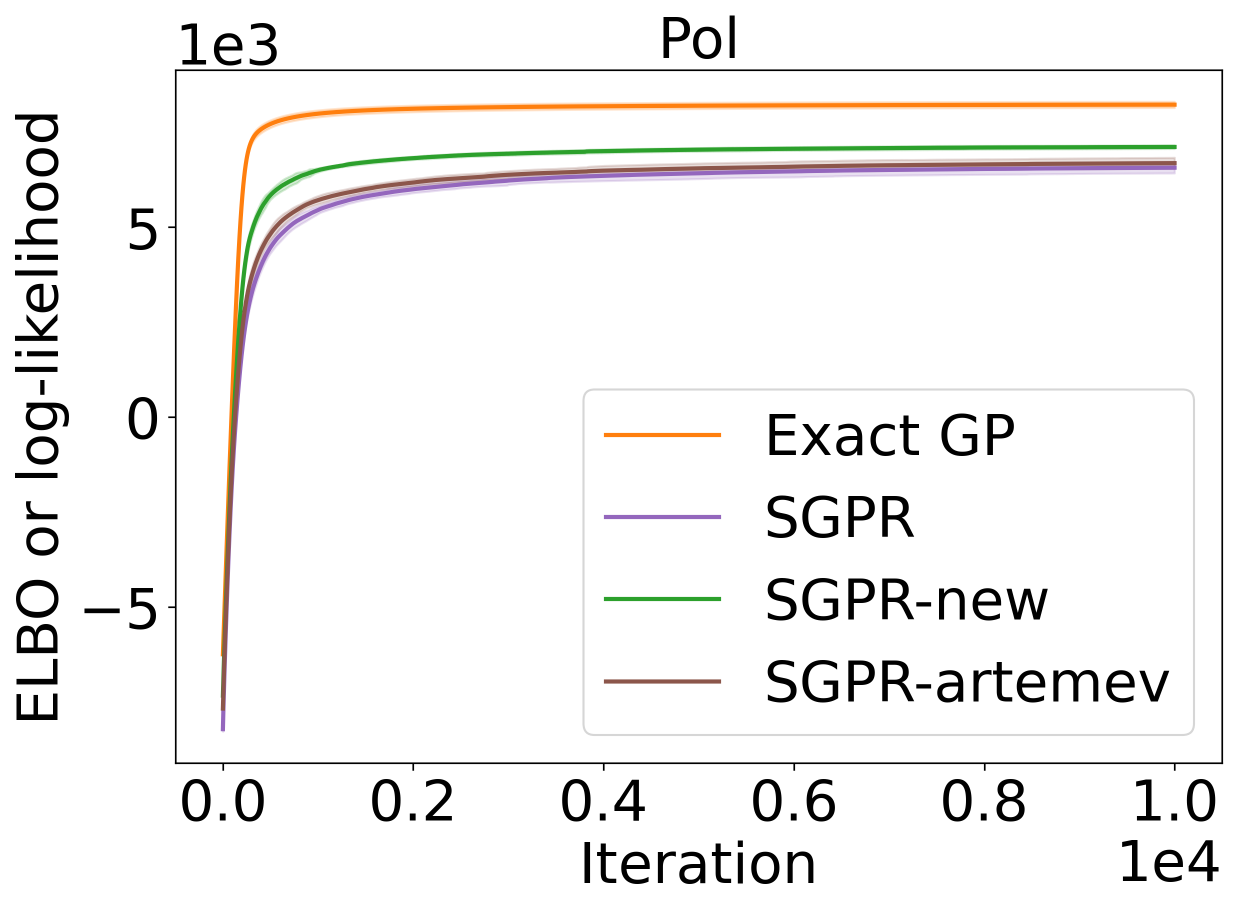} &
\includegraphics[scale=0.25]
{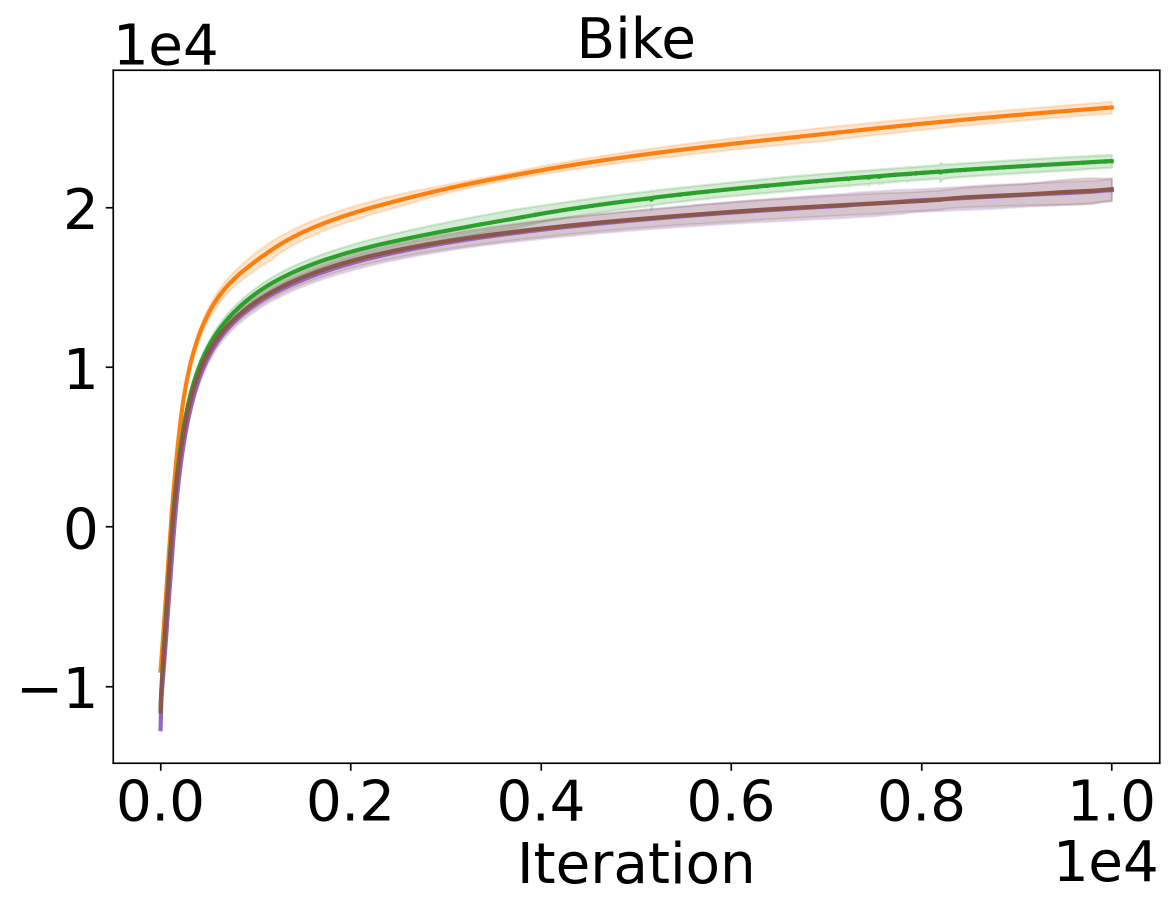} &
\includegraphics[scale=0.25]
{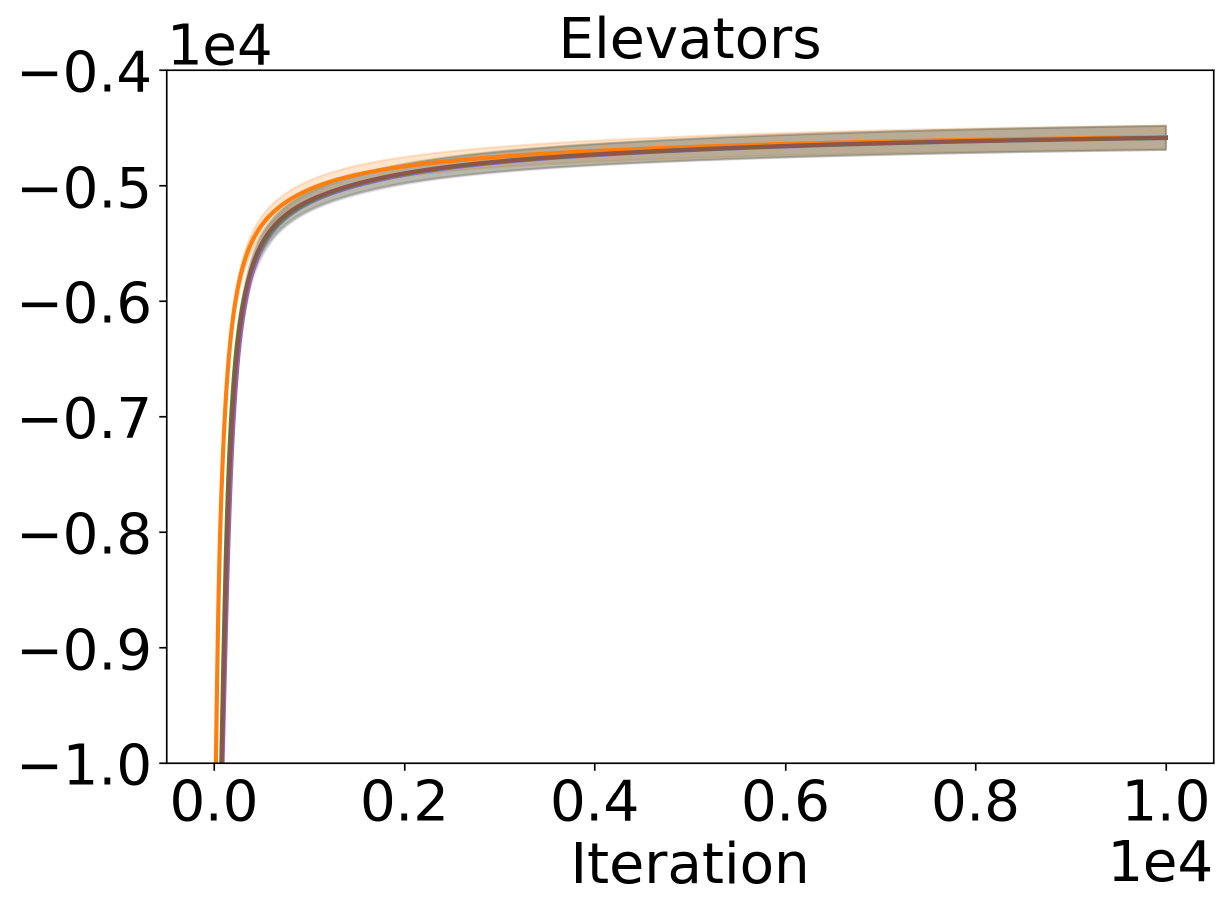} \\
\includegraphics[scale=0.25]
{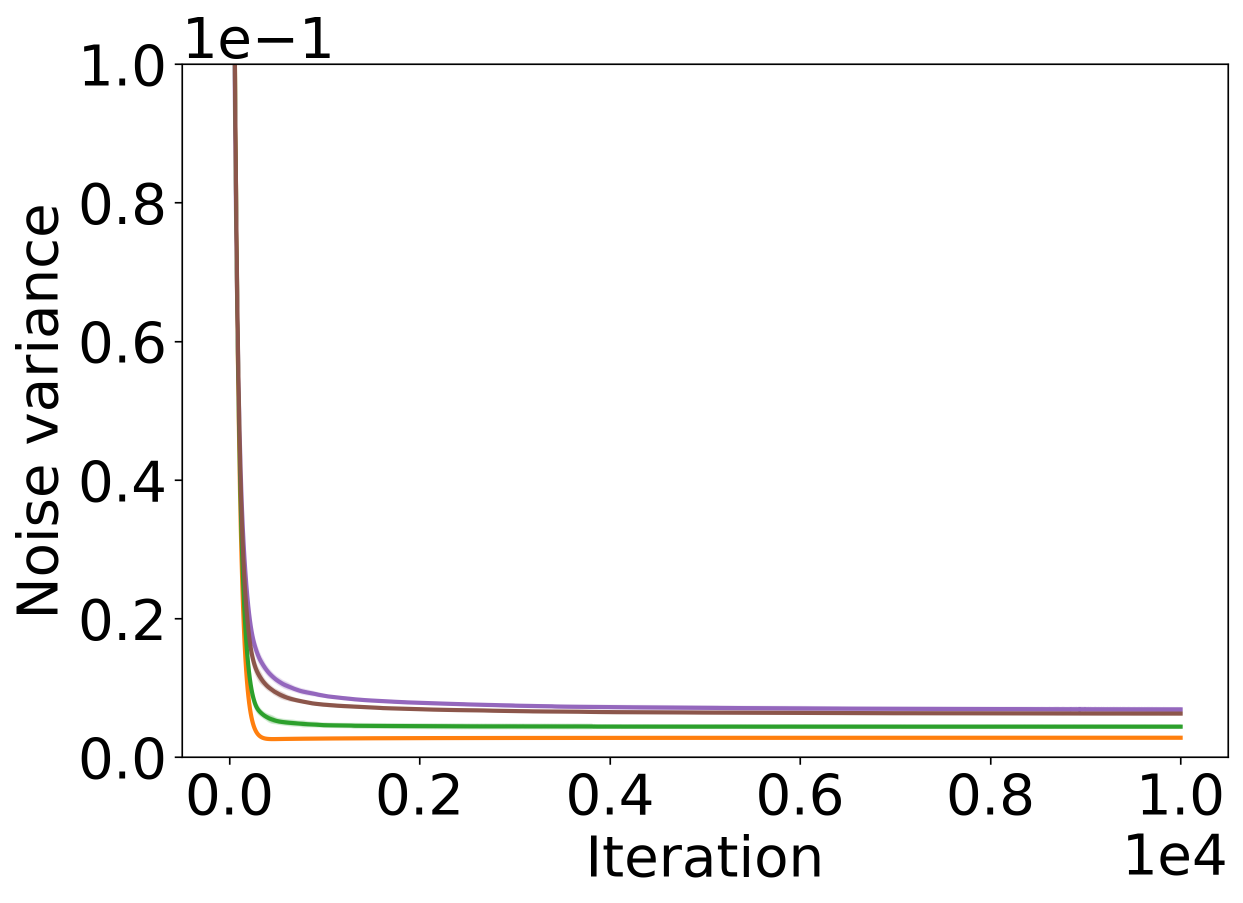} &
\includegraphics[scale=0.25]
{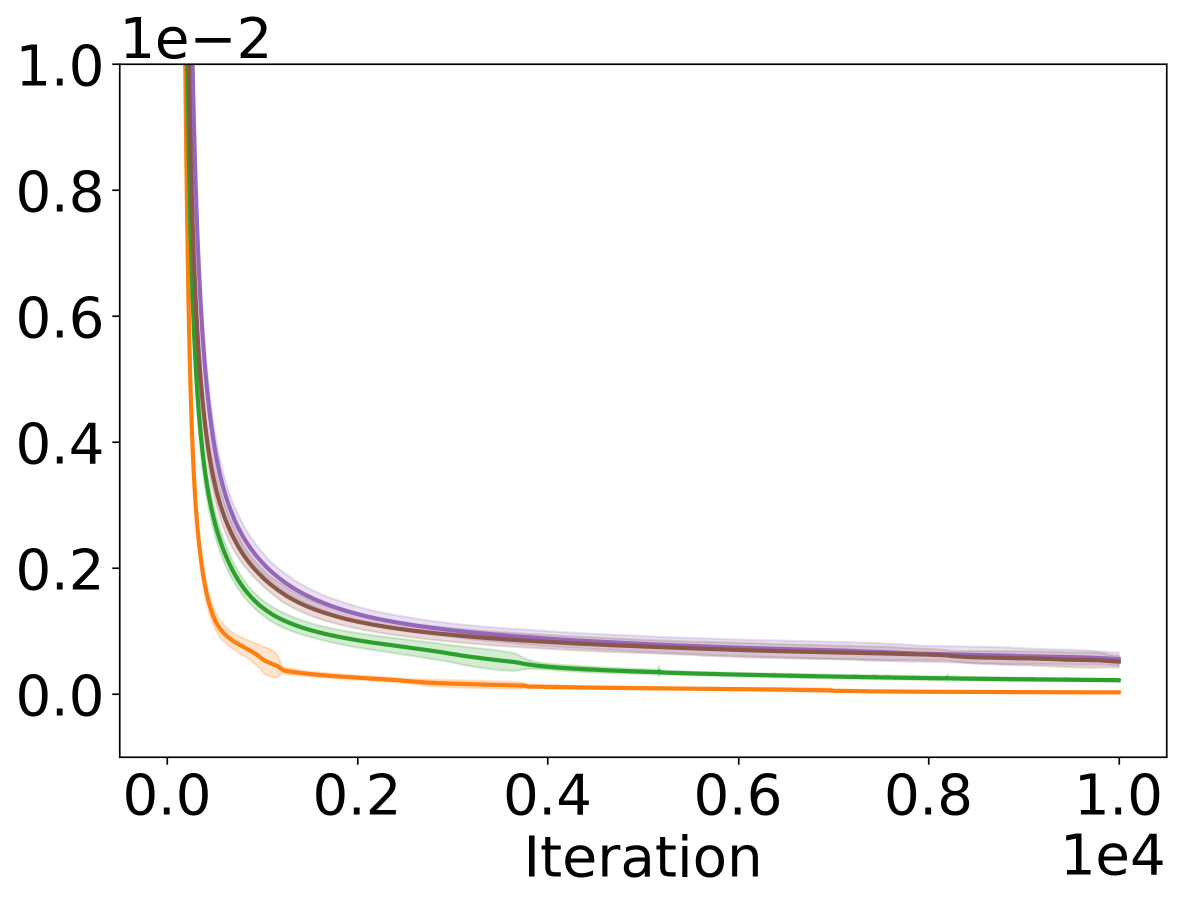} &
\includegraphics[scale=0.25]
{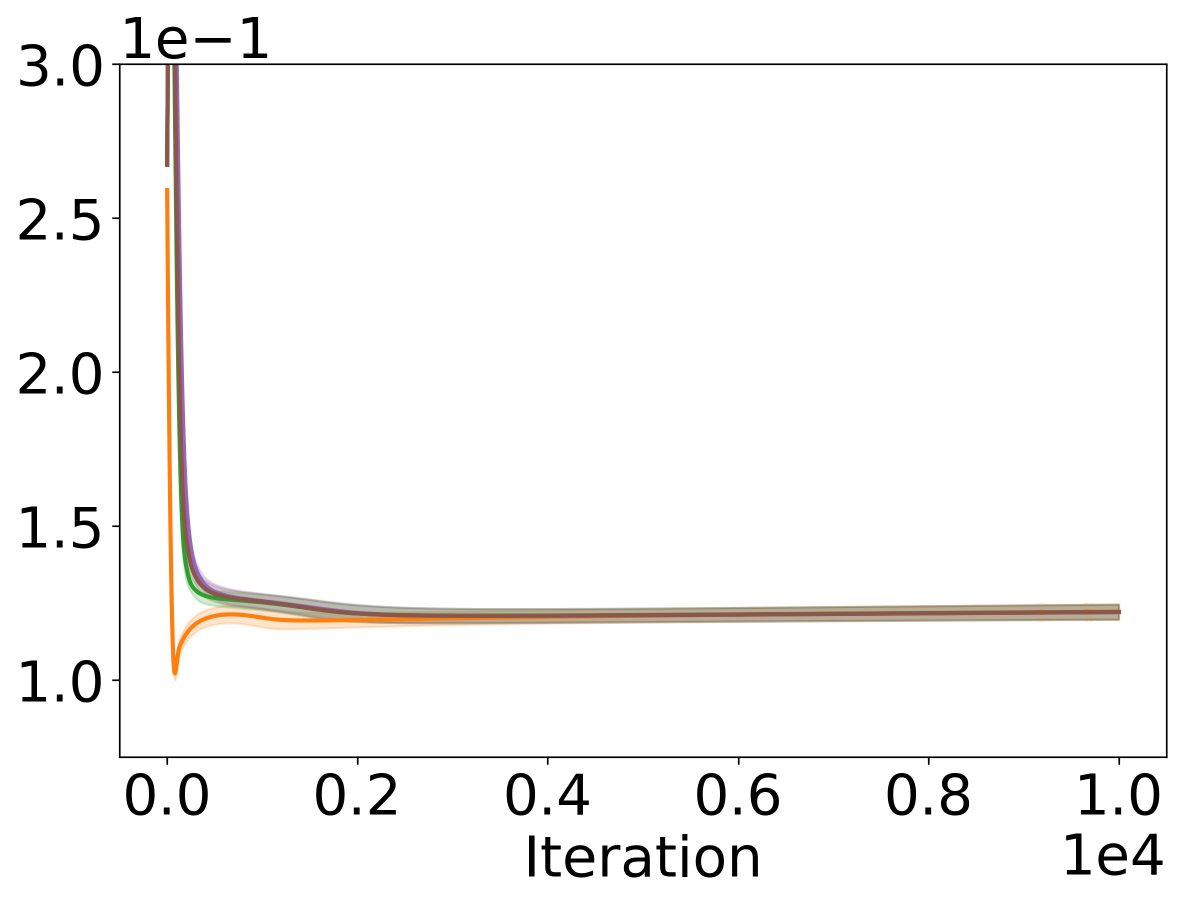} \\
\includegraphics[scale=0.24]
{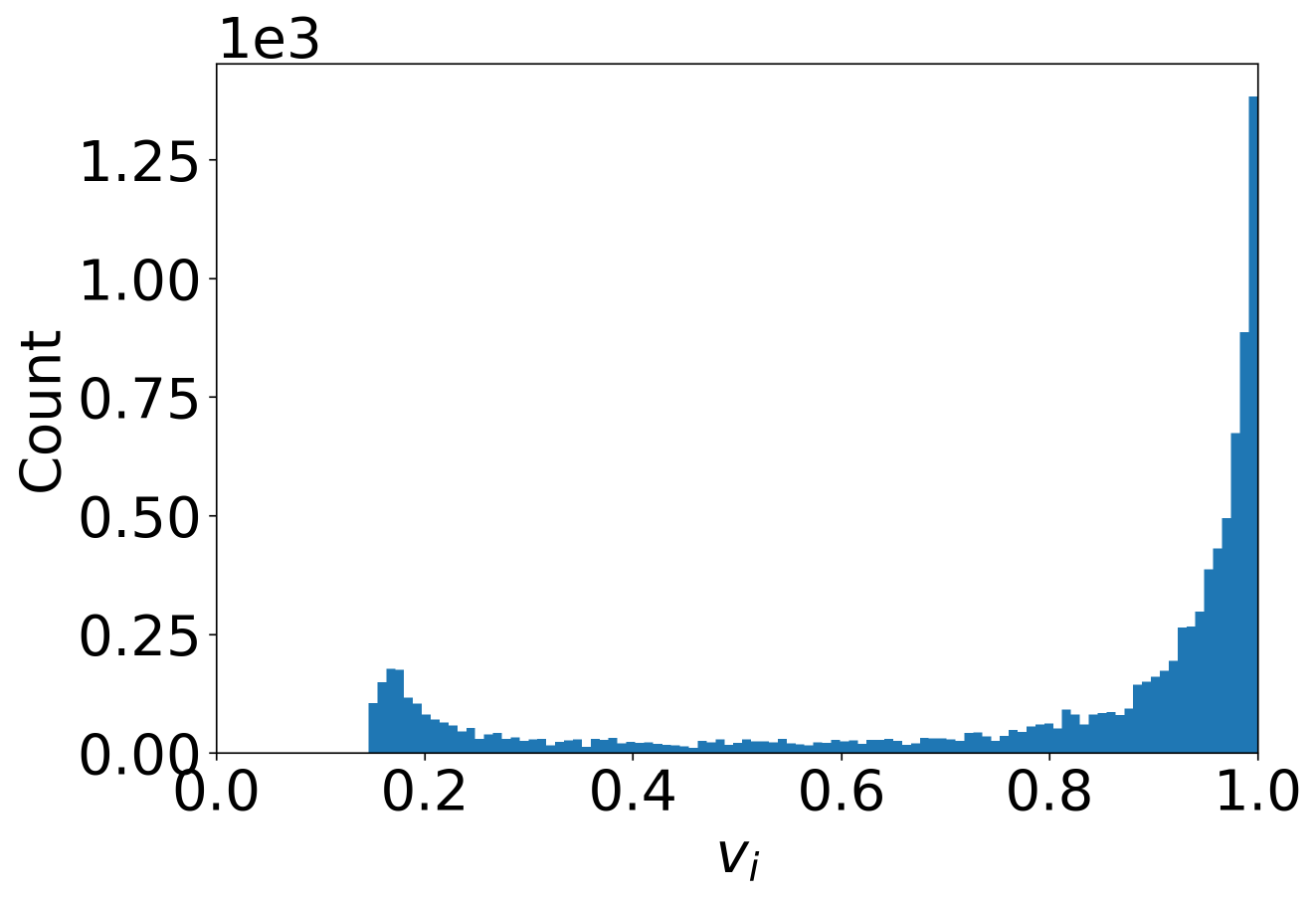} &
\includegraphics[scale=0.24]
{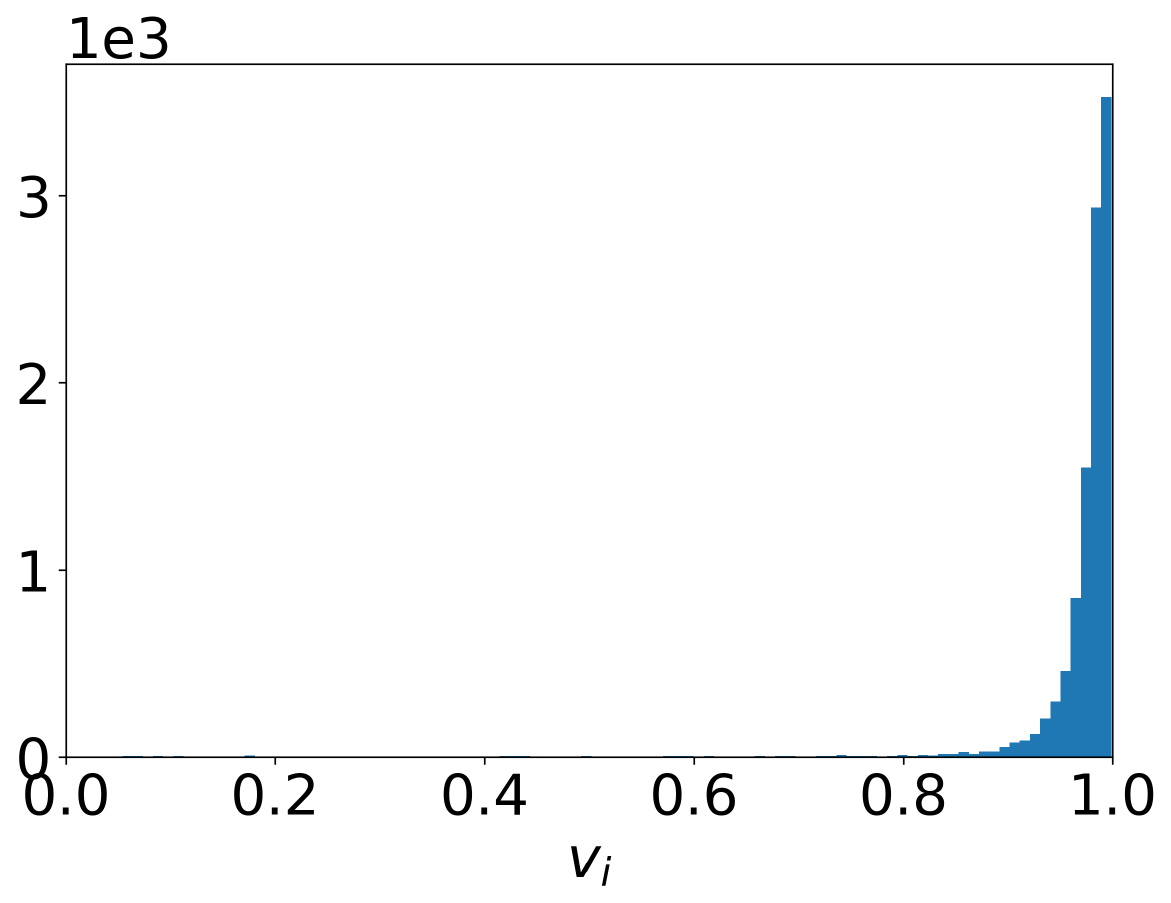} &
\includegraphics[scale=0.24]
{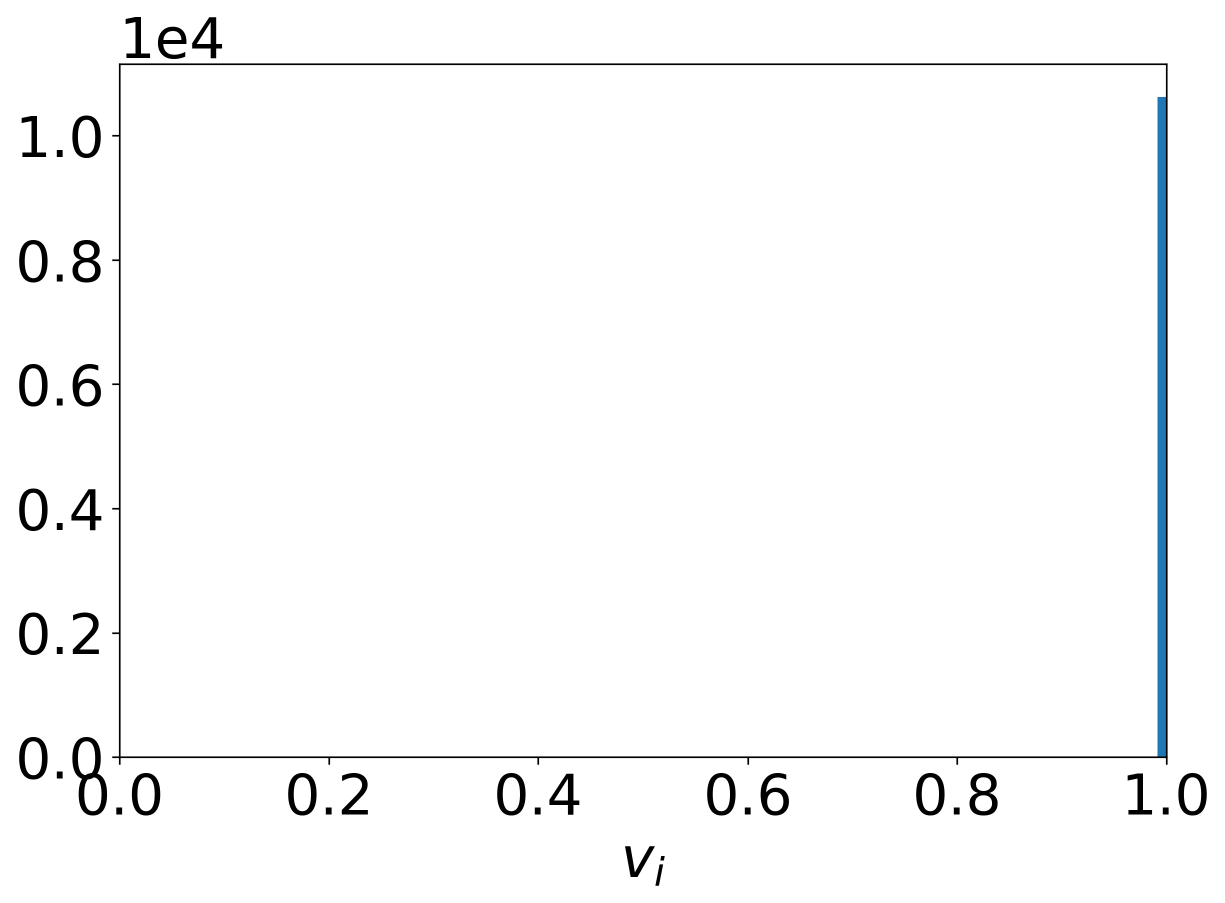}
\end{tabular}
\caption{The two plots in each column correspond to the same dataset: first row shows the ELBO (or log-likelihood)
 for all four methods (Exact GP, SGPR, SGPR-new and SGPR-artemev) with the number of iterations and the plot in the second row shows the
  corresponding values for $\sigma^2$. SGPR methods use $M=2048$ inducing points initialized by k-means. 
  For these two first lines we plot the mean and standard error
  after repeating the experiment five times with different train-test dataset splits. For one of the runs of SGPR-new, the third line shows histograms for the estimated values of the variational parameters $v_i$.  
}
\label{fig:mediumsize2048}
\end{figure*}

\Cref{fig:mediumsize2048} shows the objective function values and noise variance parameter $\sigma^2$  across iterations when the SGPR methods use $M=2048$ inducing points.  \Cref{fig:mediumsize1024} in the main paper shows the result for $M=1024$.

% Elevators 
% train (10623, 18)
% test (3320, 18)

% Bike
% train (10600, 17) this was reduced from 11122  
% test (3476, 17)

% Poll
% train (9600, 26)
% test (3000, 26)

%\begin{table}[t]
%  \caption{Test log-likelihood values for the medium size  regression datasets. The numbers in parentheses are standard errors.  The SGPR methods used $M=1024$ inducing points.}
%\label{table:smalldatasets1024}
%\vskip 0.15in
%%\begin{small}
%\begin{center}
%  \begin{sc}
%\begin{tabular}{lcccr}
%\toprule
%& Pol  & Bike & Elevators \\
%\midrule
%Exact GP & $1.089(0.011)$ & %$3.105(0.022)$ & $-0.386(0.001)$ \\
%Exact GP & $1.089(0.011)$ & %$3.105(0.022)$ & $-0.386(0.001)$  \\
%SGPR-trace & $0.821(0.008)$ & %$2.176(0.020)$ & $-0.387(0.001)$\\
%SGPR-trace & $0.958(0.008)$  & %$2.337(0.030)$ & $-0.387(0.001)$ \\
%SGPR-log & $0.920(0.006)$ & %$2.326(0.026)$  & $-0.387(0.001)$  \\
%SGPR-log & $0.998(0.008)$  & %$2.511(0.021)$ & $-0.387(0.001)$ \\
%\bottomrule
%\end{tabular}
%\end{sc}
%\end{small}
%\end{center}
%\vskip -0.1in
%\end{table}

\subsection{Large scale regression datasets
\label{app:largescaleRegress}
}

The experimental settings are chosen to match the ones from \citet{wang2019exact} and \citet{shietal2020}, where we used GPs with a Mat{\'e}rn32 kernel (with common lengthscale). Following these settings, for all datasets we train for 100 epochs using Adam with learning rate 0.01 and minibatch size 1024.

\Cref{table:largescaleRMSE} 
reports RMSE scores, while test
log likelihood scores are given in
\Cref{table:largescaleTestLL} of the main paper. 

\begin{table*}[htbp] \vskip \baselineskip
\caption{Test RMSE values of large scale regression datasets with standard errors in parentheses. Best mean values are highlighted.} %Uses random 80\% / 20\% training and test splits, repeated 5 times. }
\label{table:largescaleRMSE}
\makebox[\textwidth][c]{
\resizebox{1.02\textwidth}{!}{
\setlength\tabcolsep{2pt}
\begin{tabular}{ l l cc cc cc cc}
\toprule
& &  Kin40k & Protein &  \footnotesize KeggDirected &  KEGGU & 3dRoad & Song & Buzz & \footnotesize HouseElectric \\
\cmidrule{3-10}
& $N$ & 25,600 & 29,267 & 31,248 & 40,708 & 278,319 & 329,820 & 373,280 & 1,311,539  \\
& $d$ & 8 & 9 & 20 & 27 & 3 & 90 & 77 & 9  \\
\midrule
%\multirow{2}{*}{SVGP} & $1024$ & 0.193(0.001) & 0.630(0.004) & 0.098(0.003) & {\bf 0.123}(0.001) & 0.482(0.001) & 0.797(0.001) & 0.263(0.001) & 0.063(0.000) \\
%& $1536$ & 0.182(0.001) & 0.621(0.004) & 0.098(0.002) & {\bf 0.123}(0.001) & 0.470(0.001) & 0.797(0.001) & 0.263(0.001) & 0.063(0.000) \\
%\midrule
From \citet{shietal2020} \\
ODVGP & $1024+1024$ 
& 0.183(0.001) & 0.625(0.004) & 0.176(0.012) & 0.156(0.004) & 0.467(0.001) & 0.797(0.001) & 0.263(0.001) & 0.062(0.000) \\
& $1024+8096$  
& 0.180(0.001) & 0.618(0.004) & 0.157(0.009) & 0.157(0.004) & 0.462(0.002) & 0.797(0.001) & 0.263(0.001) & 0.062(0.000) \\
%\midrule
SOLVE-GP & $1024 + 1024$ 
& 0.172(0.001) & 0.618(0.004) & 0.095(0.002) & 0.123(0.001) & 0.464(0.001) & 0.796(0.001) & 0.261(0.001) &  0.061(0.000) \\
%\midrule
%\multirow{1}{*}{\shortstack{SVGP}}
% \\
%& $2048$ & 0.177(0.001) & {\bf 0.615}(0.004) & 0.100(0.003) & 0.124(0.001) & 0.467(0.001) & {\bf 0.796}(0.001) & 0.263(0.000) & 0.063(0.000) \\
\midrule
SVGP  & 1024 & $0.195(0.001)$ & $0.635(0.004)$ & $0.086(0.001)$ & $0.122(0.001)$ & $0.486(0.002)$ & $0.797(0.001)$ & $0.261(0.001)$ & $0.059(0.000)$ \\
      & 2048 & $0.171(0.000)$ & $0.619(0.004)$ & $0.086(0.001)$ & ${\bf 0.121}(0.001)$ & ${\bf 0.460}(0.002)$ & ${\bf 0.795}(0.001)$ & $0.260(0.001)$ & ${\bf 0.057}(0.000)$ \\
SVGP-new  & 1024 & $0.182(0.001)$ & $0.631(0.004)$ & $0.086(0.001)$ & $0.122(0.001)$ & $0.484(0.001)$ & $0.796(0.001)$ & $0.259(0.001)$ & $0.058(0.000)$ \\
     & 2048 & ${\bf 0.158}(0.000)$ & ${\bf 0.615}(0.004)$ & $0.086(0.001)$ & ${\bf 0.121}(0.001)$ & $0.461(0.002)$ & ${\bf 0.795}(0.001)$ & ${\bf 0.258}(0.000)$ & ${\bf 0.057}(0.000)$ \\
\bottomrule 
\end{tabular}
}}
\vskip \baselineskip
\end{table*}

\subsection{Poisson regression
\label{app:poisson}}

\begin{figure*}
\centering
\begin{tabular}{cc}
\includegraphics[scale=0.3]
{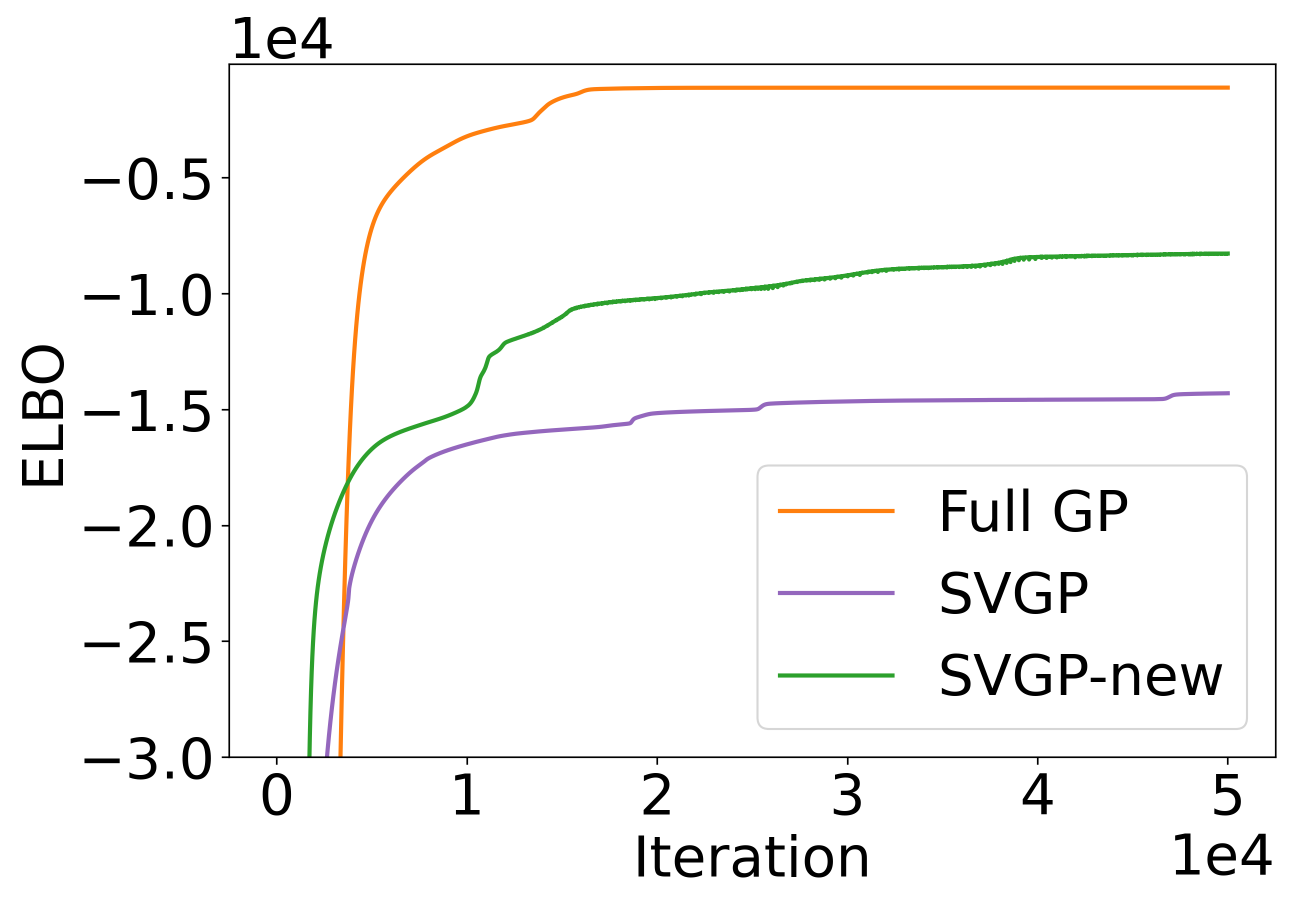} &
\includegraphics[scale=0.3]
{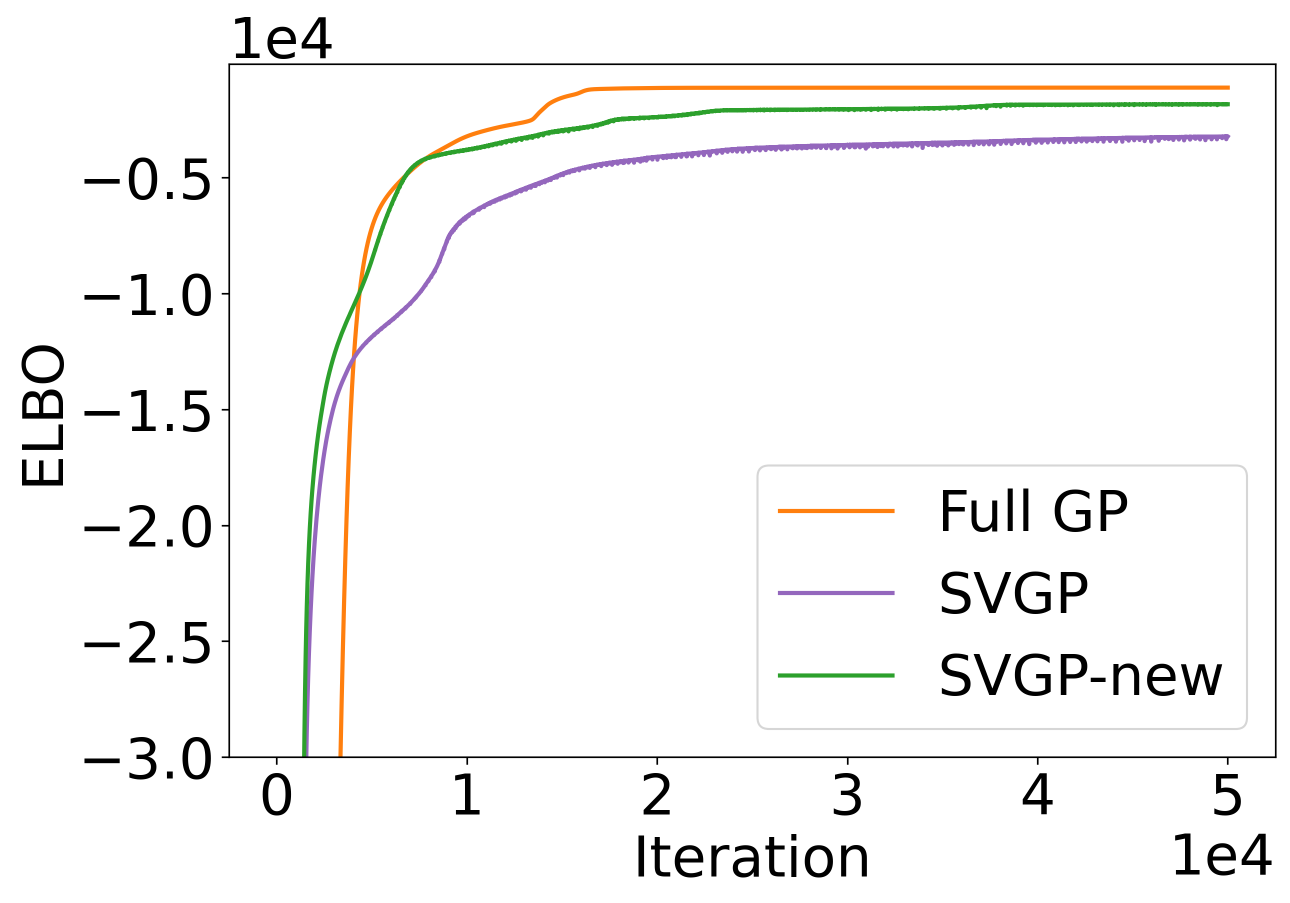} %&
%\includegraphics[scale=0.24]
%{poisson_toy_prediction_log.pdf}% \\
\end{tabular}
\caption{The % posterior 
left panel shows the lower bounds 
across iterations when the
sparse GP methods run with $M=8$ inducing points, 
while the right panel shows the corresponding plot with $M=32$ inducing points.}
\label{fig:app_poisson}
\end{figure*}

\Cref{fig:app_poisson} 
shows the ELBOs across iterations for the NYBikes 
dataset for $M=8$ and $M=32$ inducing points, while 
the plot for $M=16$ is shown in the main paper. 
\Cref{table:poisson_nybikes} presents test log-likelihood scores for the NYBikes dataset. Average values and standard errors  are computed by repeating the experiment five times
where at each repeat we randomly split the initial data into $90\%$ for training and $10\%$ for test.

%\begin{figure*}
%\centering
%\begin{tabular}{ccc}
%\includegraphics[scale=0.24]
%{poisson_toy_prediction_exact.pdf} &
%\includegraphics[scale=0.24]
%{poisson_toy_prediction_trace.pdf} &
%\includegraphics[scale=0.24]
%{poisson_toy_prediction_log.pdf} \\
%(a) & (b) & (c)
%\end{tabular}
%\caption{({\bf left}) %shows the % posterior 
%predictions (means with 2-standard deviations) over counts (black dots) in the artificial data example  after  fitting the Full GP, and the two SVGPs. This plot superimposes all predictions in order to provide a comparative visualization; see \Cref{app:poisson} for individual plots. ({\bf middle})  shows the ELBO across optimization steps. ({\bf right}) shows the ELBO for the real NYBikes dataset.}
%\label{fig:poisson}
%\end{figure*}

\begin{table}[htbp] \vskip \baselineskip
\caption{Test log likelihoods on the NYBikes  Poisson regression dataset  with standard errors in parentheses. For the 
sparse methods we consider 
varying numbers of inducing points, i.e., $M=8,16,32$.
} %Uses random 80\% / 20\% training and test splits, repeated 5 times. }
\label{table:poisson_nybikes}
\centering
 \begin{tabular}{ l l cc cc cc}
\toprule
Full GP &  & $-5.061(0.010)$ \\
\midrule
SVGP & 8 & $-36.397(6.017)$ \\
SVGP & 16 & $-16.557(4.307)$ \\
%SVGP & 24 & $-10.471(0.526)$ \\
SVGP & 32 & $-8.556(0.728)$ \\
\midrule
SVGP-new & 8 & $-9.713(0.345)$ \\
SVGP-new & 16 & $-9.301(0.296)$ \\
%SVGP-new & 24 & $-9.259(0.486)$ \\
SVGP-new & 32 & $-8.203(0.190)$ \\
\bottomrule 
\end{tabular}
\vskip \baselineskip
\end{table}
Regarding the scalar value of $v$ for the toy Poisson regression the learned value was around $v=0.675$, while or the real Poisson example in NYBikes, the value gets very small below $0.01$.